\documentclass[10pt,twocolumn,letterpaper]{article}
\usepackage[pagenumbers]{cvpr} 
\usepackage{booktabs}
\usepackage{multirow}
\usepackage{array}
\usepackage{graphicx}
\usepackage{pifont}
\usepackage{cuted}
\usepackage{caption}
\usepackage[normalem]{ulem}

\newcommand{\cmark}{\textcolor{green!70!black}{\ding{51}}}
\newcommand{\xmark}{\textcolor{red!80!black}{\ding{55}}}

\newcommand{\mstd}[2]{#1{\tiny\,\(\pm\)\,#2}}

\usepackage[most]{tcolorbox}
\usepackage{enumitem}
\newtcolorbox{promptbox}[1]{
  enhanced,
  breakable,
  title={#1},
  colback=black!7,
  colbacktitle=blue!70!black,
  coltitle=white,
  fonttitle=\bfseries,
  boxrule=0pt,
  sharp corners,
  left=1.5mm,
  right=1.5mm,
  top=1mm,
  bottom=1mm
}

\lstdefinestyle{promptcode}{
  basicstyle=\ttfamily\scriptsize,
  columns=fullflexible,
  keepspaces=true,
  showstringspaces=false,
  breaklines=true,
  frame=none,
  aboveskip=2pt,
  belowskip=2pt
}

\definecolor{cvprblue}{rgb}{0.21,0.49,0.74}
\usepackage[pagebackref,breaklinks,colorlinks,allcolors=cvprblue]{hyperref}

\def\confName{CVPR}
\def\confYear{2027}

\title{ScIntBench: Physical Interaction Plausibility in 3D Scene Generation}

\author{
Wendi Ying$^{1}$ \quad Daochang Liu$^{1}$ \quad Leonid Sigal$^{2,3}$ \quad Ajmal Saeed Mian$^{1}$\\
$^{1}$The University of Western Australia \quad
$^{2}$University of British Columbia \quad
$^{3}$Canada CIFAR AI Chair\\
{\href{https://scintbench.github.io}{https://scintbench.github.io}}}

\begin{document}
\twocolumn[{%
\renewcommand\twocolumn[1][]{#1}%
\maketitle
\vspace{-8.5mm}
\begin{center}
    \includegraphics[width=0.99\textwidth]{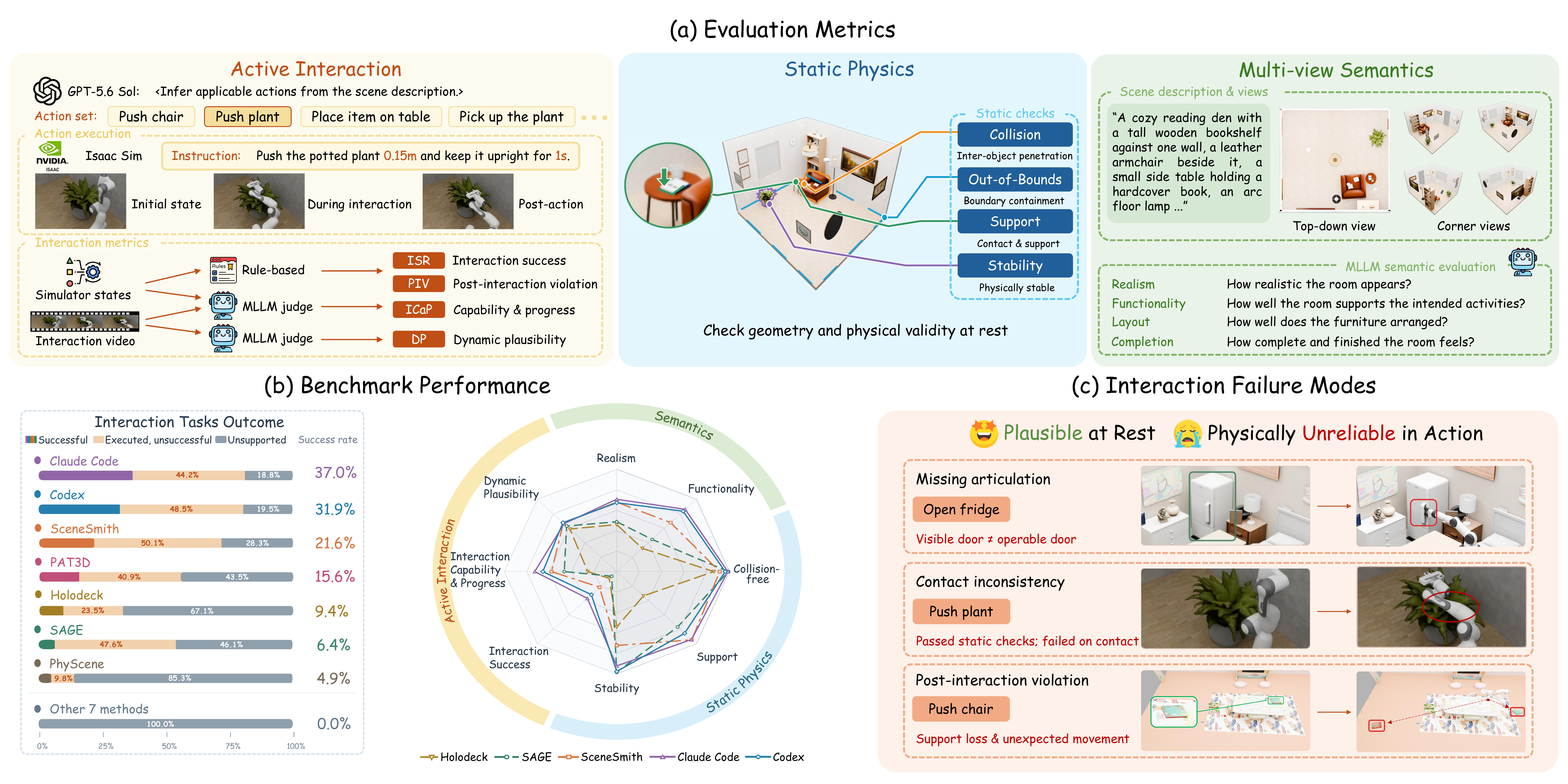}
    \captionsetup{hypcap=false}
    \captionof{figure}{\textbf{Overview of ScIntBench.} (a) ScIntBench evaluates generated 3D indoor scenes with \textbf{simulation-based active interaction metrics}, which execute scene-specific actions in a physics simulator to measure interaction success, post-interaction violations, capability progress, and dynamic plausibility. These are complemented by static physical plausibility evaluation and multi-view semantic evaluation. (b) We benchmark 14 scene generation methods, and (c) identify representative failure modes, revealing that plausible appearance at rest does not guarantee interaction readiness.}
    \label{fig:overview}
    \captionsetup{hypcap=true}
\end{center}
}]

\begin{abstract}
Text-conditioned 3D indoor scene generation is increasingly shifting toward simulation-ready, interactive environments. However, existing evaluations mainly focus on semantic fidelity and physical validity at rest, and do not sufficiently assess interaction-time physical plausibility. We introduce ScIntBench, a unified benchmark to evaluate interaction readiness through standardized physical interactions. ScIntBench consists of 50 prompts (10 room types with 5 prompts each) and metrics that measure 7 interaction categories (\eg, open, close, push, pick, place, stack). We propose new metrics to evaluate interaction success, capability and progress, post-interaction violations, and dynamic plausibility, which complement multi-view semantic and static physical evaluation. Using ScIntBench we evaluate 14 scene generation methods, analyzing a total of 8,176 generated scene-interaction instances. These extensive experiments reveal that strong semantic and static physical performance do not necessarily translate into interaction readiness: 7 of the 14 evaluated methods cannot support any prescribed interaction. These findings identify two fundamental challenges in generating interaction-ready scenes: providing the necessary physical interaction capabilities and ensuring reliable physical behavior under active manipulation. To address these challenges, we further propose an interaction-aware coding-agent method that leverages physical interaction feedback to iteratively improve generated scenes, achieving greater interaction success and physical reliability.

\end{abstract}    
\section{Introduction}
\label{sec:intro}
Text-conditioned 3D indoor scene generation has made substantial progress, enabling the automatic construction of increasingly complex environments from natural-language descriptions. Recent advances have expanded scene generation from data-driven generative models~\cite{paschalidou2021atiss,tang2024diffuscene,lin2024instructscene,yang2024physcene} to LLM/VLM-based workflows~\cite{sun2025layoutvlm,yang2024holodeck,lin2026pat3d,feng2023layoutgpt}, specialized agentic pipelines~\cite{Xia_2026_CVPR,pfaff2026scenesmith,yang2026sceneweaver}, and general-purpose coding agents operating through modeling tools~\cite{openai2026codex,anthropic2026claudecode}. This evolution has broadened the goal of scene generation beyond visual content creation toward simulation-ready environments for robotics, embodied AI, and interactive applications. However, as generation systems become increasingly capable, a fundamental question remains:\textit{ Are generated 3D scenes truly ready for physical interaction, and can existing evaluations reliably assess this capability?}

Existing evaluation protocols for generated 3D scenes primarily assess semantic quality and physical plausibility at initialization~\cite{yang2026sceneweaver,tam2026sceneeval}. Semantic evaluation examines visual fidelity while physical evaluation measures collision, boundary violations, support correctness, and passive stability under gravity. Recent works improve semantic, structural, and agent-centric evaluation of generated 3D environments~\cite{sengupta2026scenecritic,ning2026vibeworlding}, as summarized in~\cref{tab:evaluation_protocol_comparison}. However, these evaluations provide limited evidence of whether generated scenes can support physical manipulation and exhibit plausible responses to active interaction.

A scene that is geometrically valid and plausible at rest is not necessarily interaction-ready (\cref{fig:overview} c). For example, a refrigerator may appear correctly placed but lack the articulation required 
or the space required 
to open its door; a chair may remain stable under gravity yet behave implausibly when pushed or subjected to a load. Such failures expose two distinct limitations: interaction availability, which concerns whether a scene provides the prerequisites for an intended action, and interaction reliability, which concerns whether that action can be successfully executed with physically plausible outcomes. This motivates two research questions: (1) To what extent do conventional semantic and static evaluations characterize interaction readiness? (2) What limits interaction readiness across different scene generation paradigms: missing capabilities, unsuccessful execution, or implausible physical responses?

To address these limitations and advance simulation-ready scene generation, we present \textbf{ScIntBench}, a comprehensive benchmark designed to systematically evaluate the interaction readiness of generated 3D scenes across diverse generation paradigms, together with their semantic fidelity and static physical validity. ScIntBench has three evaluation dimensions: (1) \textit{active interaction evaluation}, (2) \textit{multi-view semantic evaluation}, and (3) \textit{static physical plausibility evaluation}. Specifically, the first uses standardized physics-based interaction tests that assess action executability, interaction progress, post-interaction physical integrity, and dynamic plausibility. The second uses a top-down and four corner views to better capture spatial arrangements and object relationships, while the third employs geometric checks and passive stability simulation. To ensure fair comparisons, we evaluate scenes as generated, without adding missing articulations or physical properties. Together, these evaluations characterize not only the initial configuration of a generated scene but also its behavior during physical interaction, as illustrated in~\cref{fig:overview}.

We benchmark 14 state-of-the-art 3D scene generation systems across 10 room types and 7 interaction task categories under the same text prompts to systematically investigate their interaction readiness. Our evaluation reveals substantial interaction-time failures that are overlooked by conventional initialization-time assessments. Despite achieving strong semantic fidelity and static physical validity, several methods generate scenes that frequently fail under active interaction due to missing articulations, inconsistent physical metadata, or post-interaction physical violations. By distinguishing interaction availability from execution reliability, our analysis reveals two complementary barriers to simulation readiness: providing the necessary interaction capabilities and ensuring physically reliable behavior when those capabilities are exercised.

To address the identified limitations, we further propose an interaction-aware refinement method that leverages physical interaction feedback to improve generated scenes. Specifically, it employs a physics critic to identify static and interaction-time physical failures, guiding iterative scene refinement. Experiments demonstrate that this feedback-driven refinement improves interaction success and physical reliability under our benchmark.

In summary, our main {\bf contributions} are fourfold:
\begin{itemize}
    \item We introduce ScIntBench, a unified benchmark for evaluating the interaction readiness of generated 3D indoor scenes. It contains 584 interaction tasks and establishes a systematic protocol for incorporating new scene descriptions, interaction tasks, and generation methods.

    \item We propose 4 interaction-based metrics to characterize simulation readiness beyond physical validity at rest, measuring interaction success, capability progress, post-interaction violations, and dynamic plausibility.

    \item We systematically benchmark 14 representative scene generation methods across 10 room types and 7 interaction categories, yielding 8,176 evaluation instances, highlighting key challenges between static scene quality and interaction readiness, and identifying key failure modes.

    \item We propose an interaction-aware refinement method that leverages static and interaction-time physical feedback to iteratively improve generated scenes, achieving greater physical reliability under active interactions.
\end{itemize}

\begin{table*}[t]
\centering
\caption{Comparison with related benchmarks and protocols. \textbf{Semantic view setting}: viewpoints for semantic evaluation. \textbf{Static physics}: geometric and physical checks. \textbf{Active interaction physics}: evaluates physical behavior during manipulation. \textbf{Methods Evaluated}: TSG (task-specific generators), LVG (LLM/VLM-based generators), SAG (specialized agentic systems), GCA (general-purpose coding agents).}
\label{tab:evaluation_protocol_comparison}

\resizebox{\textwidth}{!}{
\begin{tabular}{
l l c c
*{4}{>{\centering\arraybackslash}p{0.65cm}}
}
\toprule
\multirow{2}{*}{\textbf{Evaluation / Benchmark}}
& \multirow{2}{*}{\textbf{Semantic View Setting}}
& \multirow{2}{*}{\textbf{Static Physics}}
& \multirow{2}{*}{\textbf{Active Interaction Physics}}
& \multicolumn{4}{c}{\textbf{Methods Evaluated}}\\
\cmidrule(lr){5-8}
&
&
&
& {\footnotesize\textbf{TSG}}
& {\footnotesize\textbf{LVG}}
& {\footnotesize\textbf{SAG}}
& {\footnotesize\textbf{GCA}}\\
\midrule

SceneWeaver~\cite{yang2026sceneweaver}
& Top-down
& \cmark
& \xmark
& \cmark
& \cmark
& \cmark
& \xmark\\

SceneEval~\cite{tam2026sceneeval}
& Top-down
& \cmark
& \xmark
& \cmark
& \cmark
& \xmark
& \xmark\\

SceneCritic~\cite{sengupta2026scenecritic}
& Per-view analysis
& \cmark
& \xmark
& \xmark
& \cmark
& \xmark
& \xmark\\

LEGO-Eval~\cite{kang2026apples}
& Constraint-specific
& \xmark
& \xmark
& \xmark
& \cmark
& \xmark
& \xmark\\

VWE-Bench~\cite{ning2026vibeworlding}
& 5 views
& \cmark
& \xmark
& \xmark
& \xmark
& \cmark
& \cmark\\

3DTindo-Bench~\cite{qi2026global}
& Top-down + side
& \cmark
& \xmark
& \xmark
& \cmark
& \xmark
& \xmark\\

AuthorBench~\cite{xu2026muse}
& Scene state + renders
& \cmark
& \xmark
& \xmark
& \cmark
& \cmark
& \xmark\\

Code4Scene~\cite{ye2026code4scene}
& 4 overview + targeted
& \cmark
& \xmark
& \xmark
& \xmark
& \xmark
& \cmark\\
\midrule
\textbf{ScIntBench (Ours)}
& Top-down + 4 corner
& \cmark
& \cmark
& \cmark
& \cmark
& \cmark
& \cmark\\

\bottomrule
\end{tabular}
}
\end{table*}

\section{Related Work}
\label{sec:related}
\noindent {\bf 3D Indoor Scene Generation.} Early data-driven approaches learn object distributions and spatial arrangements from large-scale indoor datasets~\cite{paschalidou2021atiss,tang2024diffuscene,lin2024instructscene} such as 3D-FRONT~\cite{fu20213d}. To enable more flexible text-conditioned generation, subsequent methods incorporate semantic representations, vision-language models (VLMs), and spatial reasoning for scene planning and layout generation~\cite{feng2023layoutgpt,sun2025layoutvlm,yang2024holodeck,pun2026hsm}. More recently, agentic frameworks have extended scene synthesis through multi-step planning, tool use, and iterative refinement~\cite{yang2026sceneweaver,Xia_2026_CVPR,pfaff2026scenesmith}. Additionally, general-purpose coding agents have demonstrated the ability to construct and modify 3D worlds through programmatic tools~\cite{ning2026vibeworlding,ye2026code4scene}. This progression has diversified scene generation paradigms, motivating unified evaluation across systems with different generation mechanisms and capabilities.

\vspace{1mm}
\noindent {\bf Simulation-Ready and Agentic Scene Generation.} 
There is a growing research trend toward generating physically plausible and interactive environments for embodied AI through geometric constraints and physics-guided optimization~\cite{yang2024physcene,lin2026pat3d}. Agentic frameworks further enable tool-based scene construction, executable interaction representations, and iterative refinement~\cite{wang2026scenecode,xiao2026beyond,yang2026sceneweaver,xu2026muse,zhang2026scenepilot,ran2026scenemosaic,ling2026scenethesis,wang2026physcensis}. SAGE~\cite{Xia_2026_CVPR} incorporates visual and physics critics to iteratively refine scenes through simulation-based validation. SceneSmith~\cite{pfaff2026scenesmith} integrates articulated assets and physical property estimation to construct simulation-ready environments for downstream robotic tasks. EmbodiedSmith~\cite{qin2026embodiedsmith} further leverages simulated task execution feedback to jointly refine scenes and task plans. Complementary approaches explore efficient tool orchestration, code-based generation, and functional or task-driven scene refinement~\cite{he2026sceneorchestra,yang2026code,feng2026funcroom,xu2026gif}. Despite these advances, scene refinement often focuses on initialization-time physical validity or downstream task success, rather than explicitly addressing physical violations during active interaction.

\vspace{1mm}
\noindent {\bf Evaluation of Text-Conditioned 3D Scene Generation.} 
Existing evaluations primarily assess semantic fidelity, geometric consistency, and static physical plausibility through visual assessments and constraint-based checks~\cite{yang2026sceneweaver,tam2026sceneeval,sengupta2026scenecritic,kang2026apples,qi2026global}. More recent benchmarks extend the evaluation to agentic and general-purpose 3D scene construction and editing, assessing instruction satisfaction, geometric validity, and scene-state preservation~\cite{xu2026muse,ning2026vibeworlding,ye2026code4scene}. SceneTeract~\cite{maillard2026sceneteract} evaluates agent-specific activity feasibility through navigability, reachability, and interaction clearance. SceneSmith~\cite{pfaff2026scenesmith} and EmbodiedSmith~\cite{qin2026embodiedsmith} further incorporate simulated robotic manipulation, but their execution-based evaluations primarily focus on downstream task success rather than scene-level physical reliability. Existing approaches lack a unified protocol for evaluating interaction availability and physical reliability during and after active object manipulation across diverse scene generation paradigms. ScIntBench addresses this gap through multi-view semantic evaluation, static physical evaluation, and standardized simulation-based interaction tests (\cref{tab:evaluation_protocol_comparison}).
\section{ScIntBench}
\label{sec:bench}
We present ScIntBench, a unified framework for evaluating interaction readiness of text-conditioned 3D indoor scenes. Our evaluation consists of semantic fidelity, static physical validity, and interaction-time physical reliability (\cref{fig:overview} a). Metric details are provided in Supplementary~\cref{sec:metrics_detail}.

\subsection{Benchmark Construction}
ScIntBench builds on the text prompts of SceneEval~\cite{tam2026sceneeval}, covering 10 indoor room types, and extends them with standardized interaction specifications. For each room description, an LLM proposes candidate interactions on relevant object categories. Each specification defines a target object, an intended action, the required interaction capabilities, success conditions, and applicable physical constraints. These specifications are constructed independently of the generated scenes and shared across all methods (more details of construction can be found in Supplementary~\cref{sec:construction}).

\subsection{Active Interaction Evaluation}
A scene that satisfies semantic and static physical constraints may still lack the capabilities required for active interaction. For prompt $p$, let $S_{m,p}$ denote the scene generated by method $m$ and $\mathcal{T}_p=\{t_1,\ldots,t_{N_p}\}$ the prescribed interactions. For each $t\in\mathcal{T}_p$, we first verify the target object and its native physical capabilities, then execute supported actions in a physics simulator. Execution coverage (\textbf{EXE}) reports the fraction of prescribed tasks that are executed. Unsupported tasks remain included in ISR and ICaP, but are excluded from PIV and DP, which are marked N.S.A.\ for methods without any executed task. We evaluate the outcomes using four complementary metrics.

\vspace{1mm}
\noindent {\bf Interaction Success Rate (ISR).} ISR measures the proportion of tasks in $\mathcal{T}_p$ whose goals are achieved, based on simulator-observable state and action-specific success criteria. Unresolved controller or simulation failures are flagged separately rather than attributed to the generated scene. 

\vspace{1mm}
\noindent {\bf Interaction Capability Progress (ICaP).} ICaP captures the progress that binary success conceals, using a six-level rubric: Target Absent (0), Target Present (2), Physically Interactable (4), Interaction Initiated (6), Substantial Progress (8), and Task Success (10). An MLLM assigns the level from execution evidence, and ICaP is averaged over $\mathcal{T}_p$.

\vspace{1mm}
\noindent {\bf Post-Interaction Physical Violation (PIV).} For each executed interaction $t \in \mathcal{T}_p$, PIV measures the percentage of scene objects that exhibit physical violations after a standardized settlement period. Violations include unintended displacement, new or worsened interpenetration, out-of-bounds placement, and loss of required support. Intended object motion is not counted.

\vspace{1mm}
\noindent {\bf Dynamic Plausibility (DP).} For the executed interaction $t \in \mathcal{T}_p$, DP provides complementary evaluation of physical behavior during interaction. An MLLM scores the rendered interaction video on a 0--10 scale, covering motion naturalness, contact, gravity, support, friction, and articulation consistency. The rubric separates observable violations from insufficient visual evidence, and task failure alone does not imply implausible motion.

\subsection{Semantic and Static Physical Evaluation}
To complement the evaluation of interaction readiness, we also evaluate semantic quality and static physical validity by adapting existing evaluation protocols with modifications.

\vspace{1mm}
\noindent {\bf Multi-view Semantic Evaluation.} Each scene is rendered from a top-down and four corner views to capture the global layout while reducing occlusion and reveal spatial relationships that may be ambiguous from a single viewpoint.  Following~\cite{yang2026sceneweaver}, an MLLM jointly scores all views against the corresponding text prompt for realism (\textbf{Real}), functionality (\textbf{Func}), layout (\textbf{Lay}), and completeness (\textbf{Comp}).

\vspace{1mm}
\noindent {\bf Static Physical Evaluation.} We evaluate static physical validity through geometric checks and passive stability simulation. Following~\cite{tam2026sceneeval,Xia_2026_CVPR}, geometric checks measure collisions (\textbf{COL}), support (\textbf{SUP}), and out-of-bounds placement (\textbf{OB}). Passive stability (\textbf{STAB}) simulates the scene under gravity and detects unintended movement or loss of support; non-simulatable scenes are marked N.S.A.

\subsection{Standardized Evaluation Protocol}
All scenes are evaluated with their original geometry, object poses, and native physical properties. Missing physical parameters or articulations are not supplemented. Scenes are rendered in Blender~\cite{blender} and simulated in Isaac Sim~\cite{nvidiaisaacsim}, and conversion accuracy is verified by comparing geometry and physical properties before and after conversion (more details in Supplementary~\cref{sec:simulation_detail}). The action executors and the success criteria are calibrated in the reference simulations before evaluation, and identical execution parameters are used across the methods.

\section{Experiments}
\label{sec:experi}
\begin{table*}[htbp!]
\centering
\caption{{\bf Performance comparison of different scene generation methods on ScIntBench.} Metrics: Real = realism, Func = functionality, Lay = layout, Comp = completeness, COL = collision rate, OB = out-of-bounds, SUP = support, STAB = stability, ISR = interaction success rate, ICaP = interaction capability progress, PIV = post-interaction violation rate, DP = dynamic plausibility. COL, OB, SUP, STAB, ISR, PIV are percentages, while all other metrics are scored on a 0--10 scale. N.S.A. denotes not support interaction due to missing targets or native physical prerequisites. We report mean values $\pm$ standard deviation across scenes. EXE is the coverage of interaction tasks executed in simulation (out of 584), PIV and DP are computed on these executed tasks only. $\uparrow$ indicates higher is better, and $\downarrow$ indicates lower is better. \textbf{Bold} = best, \underline{underline} = second best.}
\label{tab:overall_evaluation}
\setlength{\tabcolsep}{3pt}
\renewcommand{\arraystretch}{1.15}
\resizebox{\linewidth}{!}{%
\begin{tabular}{l*{8}{c}>{\color{gray}}c*{4}{c}}
\toprule
\multirow{2}{*}{Method} &
\multicolumn{4}{c}{Semantic Quality} &
\multicolumn{4}{c}{Static Physical Plausibility} &
\multicolumn{5}{c}{Interactive Physical Plausibility} \\
\cmidrule(lr){2-5}
\cmidrule(lr){6-9}
\cmidrule(lr){10-14}
&
Real $\uparrow$ &
Func $\uparrow$ &
Lay $\uparrow$ &
Comp $\uparrow$ &
COL $\downarrow$ &
OB $\downarrow$ &
SUP $\uparrow$ &
STAB $\uparrow$ &
\multicolumn{1}{c}{EXE} $\uparrow$&
ISR $\uparrow$ &
ICaP $\uparrow$ &
PIV $\downarrow$ &
DP $\uparrow$ \\
\midrule

\multicolumn{14}{l}{
    \textit{Task-specific generators}} \\

ATISS
    & \mstd{3.00}{0.99}
    & \mstd{1.20}{1.34}
    & \mstd{1.92}{1.03}
    & \mstd{1.78}{0.82}
    & \mstd{60.88}{22.38}
    & \mstd{27.45}{14.57}
    & \mstd{27.89}{16.35}
    & N.S.A.
    & 0/584
    & \mstd{0.00}{0.00}
    & \mstd{0.45}{0.57}
    & N.S.A.
    & N.S.A. \\

DiffuScene
    & \mstd{3.85}{1.04}
    & \mstd{1.10}{1.21}
    & \mstd{2.50}{0.83}
    & \mstd{2.00}{0.73}
    & \mstd{39.15}{23.06}
    & \mstd{15.56}{11.99}
    & \mstd{66.36}{19.28}
    & N.S.A.
    & 0/584
    & \mstd{0.00}{0.00}
    & \mstd{0.34}{0.42}
    & N.S.A.
    & N.S.A. \\

InstructScene
    & \mstd{3.00}{0.65}
    & \mstd{1.05}{1.36}
    & \mstd{1.95}{0.76}
    & \mstd{1.65}{0.67}
    & \mstd{35.11}{22.30}
    & \mstd{19.04}{16.77}
    & \mstd{63.68}{17.28}
    & N.S.A.
    & 0/584
    & \mstd{0.00}{0.00}
    & \mstd{0.33}{0.47}
    & N.S.A.
    & N.S.A. \\

PhyScene
    & \mstd{2.56}{0.95}
    & \mstd{1.20}{1.21}
    & \mstd{1.82}{0.87}
    & \mstd{1.46}{0.81}
    & \mstd{41.06}{21.98}
    & \mstd{39.04}{16.75}
    & \mstd{22.55}{14.86}
    & \mstd{94.66}{11.77}
    & 86/584
    & \mstd{4.89}{6.59}
    & \mstd{1.20}{1.47}
    & \underline{\mstd{2.60}{6.76}}
    & \mstd{5.40}{0.27} \\

\midrule
\multicolumn{14}{l}{
    \textit{LLM/VLM-based generators}} \\

LayoutGPT
    & \mstd{2.52}{1.05}
    & \mstd{3.22}{1.40}
    & \mstd{2.16}{1.18}
    & \mstd{1.70}{0.81}
    & \mstd{29.14}{30.58}
    & \mstd{41.37}{28.11}
    & \mstd{9.66}{16.04}
    & N.S.A.
    & 0/584
    & \mstd{0.00}{0.00}
    & \mstd{1.39}{0.67}
    & N.S.A.
    & N.S.A.\\

LayoutVLM
    & \mstd{3.04}{0.95}
    & \mstd{3.94}{1.19}
    & \mstd{3.04}{1.26}
    & \mstd{2.10}{0.74}
    & \mstd{33.63}{27.65}
    & \mstd{17.78}{19.37}
    & \mstd{76.20}{21.48}
    & N.S.A.
    & 0/584
    & \mstd{0.00}{0.00}
    & \mstd{1.66}{0.49}
    & N.S.A.
    & N.S.A.\\

Holodeck
    & \mstd{4.60}{1.14}
    & \mstd{3.22}{1.57}
    & \mstd{2.64}{0.83}
    & \mstd{2.64}{1.24}
    & \mstd{13.48}{13.38}
    & \mstd{4.95}{6.16}
    & \mstd{33.61}{15.62}
    & \mstd{54.25}{41.58}
    & 192/584
    & \mstd{9.41}{10.52}
    & \mstd{2.58}{1.58}
    & \mstd{5.95}{8.49}
    & \mstd{5.89}{1.02}\\

HSM
    & \mstd{4.20}{0.93}
    & \mstd{4.35}{1.33}
    & \mstd{3.61}{1.10}
    & \mstd{3.55}{0.91}
    & \mstd{16.30}{14.47}
    & \mstd{10.36}{8.91}
    & \mstd{73.64}{15.73}
    & N.S.A.
    & 0/584
    & \mstd{0.00}{0.00}
    & \mstd{1.62}{0.48}
    & N.S.A.
    & N.S.A.\\

PAT3D
    & \mstd{2.53}{1.01}
    & \mstd{3.60}{1.25}
    & \mstd{2.00}{1.29}
    & \mstd{2.37}{0.61}
    & \mstd{28.60}{23.64}
    & \mstd{46.89}{27.04}
    & \mstd{38.38}{25.81}
    & \mstd{60.79}{28.92}
    & 330/584
    & \mstd{15.63}{15.54}
    & \mstd{4.77}{1.52}
    & \mstd{29.28}{15.93}
    & \mstd{5.61}{0.98} \\

\midrule
\multicolumn{14}{l}{
    \textit{Specialized agentic systems}} \\

SceneWeaver
    & \mstd{3.60}{1.63}
    & \mstd{3.93}{1.78}
    & \mstd{3.87}{1.93}
    & \mstd{2.67}{1.30}
    & \mstd{12.36}{20.02}
    & \mstd{5.14}{11.96}
    & \mstd{8.58}{13.64}
    & N.S.A.
    & 0/584
    & \mstd{0.00}{0.00}
    & \mstd{1.40}{0.63}
    & N.S.A.
    & N.S.A.\\

SAGE
    & \mstd{4.88}{1.33}
    & \mstd{4.46}{1.63}
    & \mstd{3.02}{1.24}
    & \underline{\mstd{4.74}{1.98}}
    & \textbf{\mstd{0.67}{1.40}}
    & \textbf{\mstd{0.00}{0.00}}
    & \mstd{76.63}{12.46}
    & \textbf{\mstd{98.38}{3.31}}
    & 315/584
    & \mstd{6.35}{8.92}
    & \mstd{4.67}{1.57}
    & \textbf{\mstd{0.97}{1.30}}
    & \mstd{6.31}{0.91}\\

SceneSmith
    & \underline{\mstd{6.80}{1.34}}
    & \mstd{6.78}{1.88}
    & \mstd{5.80}{1.55}
    & \textbf{\mstd{5.26}{1.65}}
    & \mstd{8.35}{8.01}
    & \mstd{4.67}{4.58}
    & \textbf{\mstd{94.44}{7.07}}
    & \mstd{72.32}{26.14}
    & 419/584
    & \mstd{21.64}{18.38}
    & \mstd{5.81}{1.45}
    & \mstd{6.29}{6.61}
    & \underline{\mstd{6.72}{0.73}}\\

\midrule
\multicolumn{14}{l}{
    \textit{General-purpose coding agents}} \\

Codex
    & \mstd{6.74}{0.80}
    & \underline{\mstd{8.36}{1.06}}
    & \underline{\mstd{7.40}{1.14}}
    & \mstd{4.46}{1.30}
    & \mstd{3.48}{10.29}
    & \mstd{4.92}{9.51}
    & \mstd{85.42}{19.05}
    & \underline{\mstd{97.75}{6.85}}
    & 470/584
    & \underline{\mstd{31.93}{14.74}}
    & \underline{\mstd{6.58}{0.95}}
    & \mstd{7.70}{9.71}
    & \textbf{\mstd{6.77}{0.51}}\\

Claude Code
    & \textbf{\mstd{7.08}{0.80}}
    & \textbf{\mstd{8.62}{1.09}}
    & \textbf{\mstd{7.62}{0.81}}
    & \mstd{4.70}{1.28}
    & \underline{\mstd{1.06}{3.68}}
    & \underline{\mstd{0.40}{2.83}}
    & \underline{\mstd{94.05}{8.88}}
    & \mstd{92.30}{21.15}
    & 474/584
    & \textbf{\mstd{36.99}{20.21}}
    & \textbf{\mstd{7.30}{1.07}}
    & \mstd{15.03}{14.73}
    & \mstd{6.63}{1.16}\\

\bottomrule
\end{tabular}%
}

\end{table*}

\begin{figure*}
    \centering
    \includegraphics[width=\textwidth]{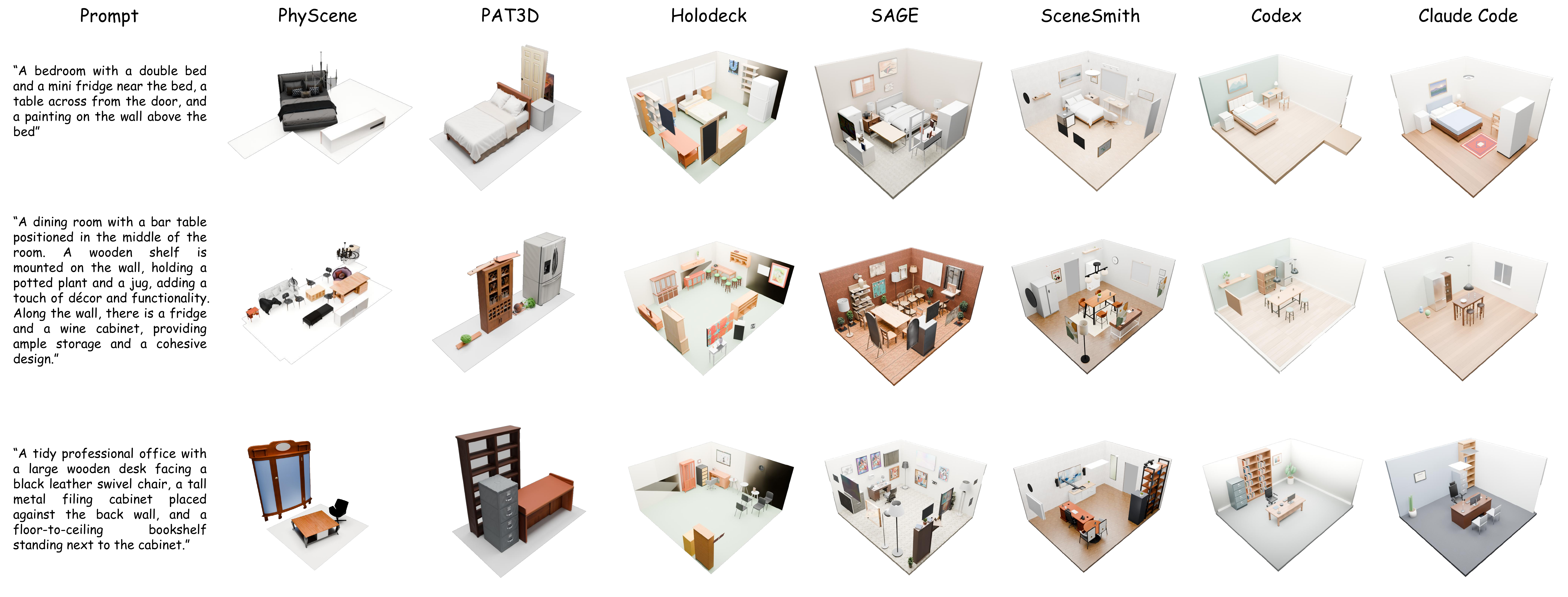}
    \captionsetup{hypcap=false}
    \captionof{figure}{\textbf{Qualitative comparison of generated scenes on ScIntBench.} We show scenes generated by PhyScene~\cite{yang2024physcene}, PAT3D~\cite{lin2026pat3d}, Holodeck~\cite{yang2024holodeck}, SAGE~\cite{Xia_2026_CVPR}, SceneSmith~\cite{pfaff2026scenesmith}, Codex~\cite{openai2026codex}, and Claude Code~\cite{anthropic2026claudecode}.}
    \label{fig:qualitative_scene}
    \captionsetup{hypcap=true}
\end{figure*}

\begin{figure*}
    \centering
    \includegraphics[width=\textwidth]{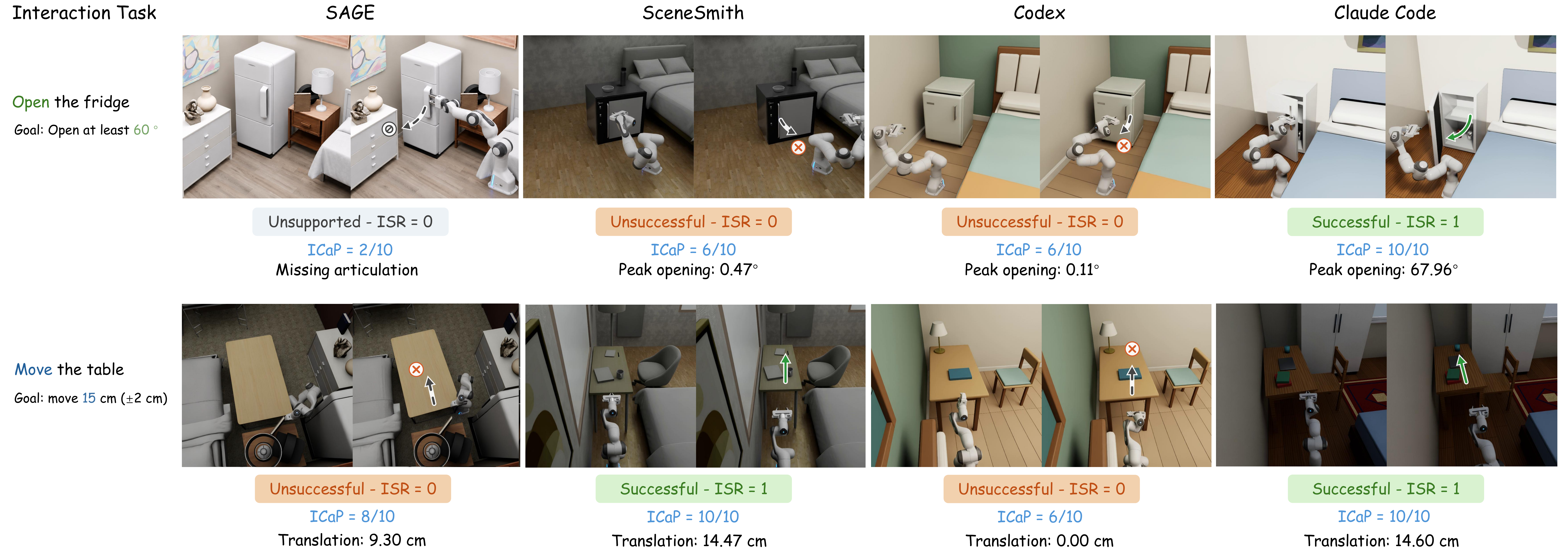}
    \captionsetup{hypcap=false}
    \captionof{figure}{\textbf{Qualitative comparison of interaction task outcomes.} We show example outcomes in scenes generated by SAGE~\cite{Xia_2026_CVPR}, SceneSmith~\cite{pfaff2026scenesmith}, Codex~\cite{openai2026codex}, and Claude Code~\cite{anthropic2026claudecode}.}
    \label{fig:qualitative_interaction}
    \captionsetup{hypcap=true}
\end{figure*}

\vspace{1mm}
\noindent {\bf Evaluated Methods.}
We evaluate 14 text-conditioned 3D indoor scene generation systems spanning four paradigms: task-specific generators, including ATISS~\cite{paschalidou2021atiss}, DiffuScene~\cite{tang2024diffuscene}, InstructScene~\cite{lin2024instructscene}, and PhyScene~\cite{yang2024physcene}; LLM/VLM-based generators, including LayoutGPT~\cite{feng2023layoutgpt}, LayoutVLM~\cite{sun2025layoutvlm}, Holodeck~\cite{yang2024holodeck}, HSM~\cite{pun2026hsm}, and PAT3D~\cite{lin2026pat3d}; specialized agentic pipelines, including SceneWeaver~\cite{yang2026sceneweaver}, SAGE~\cite{Xia_2026_CVPR}, and SceneSmith~\cite{pfaff2026scenesmith}; and general-purpose coding agents, including Codex~\cite{openai2026codex} and Claude Code~\cite{anthropic2026claudecode}. The first group learns scene distributions from 3D-FRONT, whereas the remaining methods employ language-guided planning, hierarchical scene construction, agentic tool use, or programmatic generation. 

Although ATISS and PhyScene cannot be conditioned on text descriptions, they are widely used as baselines in prior work. Following SceneEval~\cite{tam2026sceneeval}, we include them to assess the value of text conditioning; PhyScene further incorporates physics guidance, which is directly relevant to our physical and interaction metrics. Since official pretrained checkpoints are not fully available, we retrain both models on 3D-FRONT using their official implementations.

\vspace{1mm}
\noindent {\bf Benchmark Data.}
We select 50 prompts from SceneEval~\cite{tam2026sceneeval}, covering 10 types of indoor room (5 prompts each). Based on the room descriptions, we construct a total of 584 prescribed interaction tasks, averaging 12 tasks per prompt. The same task set is shared across all evaluated methods, yielding 8,176 scene-interaction evaluation instances.

\vspace{1 mm}
\noindent {\bf Implementation Details.}
We use Blender~\cite{blender} for standardized five-view rendering and Isaac Sim~\cite{nvidiaisaacsim} for physics-based simulation. GPT-5.6 is used to construct interaction tasks from room descriptions and to evaluate semantic fidelity, interaction capability progress, and dynamic plausibility. Active interactions are executed using a Franka Panda robot for contact-based manipulation and a standardized force-driven probe for surface interaction tests. Detailed baseline configurations and simulation protocols are provided in Supplementary~\cref{sec:implementation,sec:simulation_detail}.

\vspace{1mm}
\noindent {\bf Metrics.}
We report four semantic metrics (Real, Func, Lay, Comp), four static physical metrics (COL, OB, SUP, STAB), and four active interaction metrics (ISR, ICaP, PIV, DP), together with EXE. For active evaluation, ISR and ICaP are aggregated over all prescribed interactions and PIV and DP over executed interactions within each scene, then averaged equally across scenes.

\subsection{Overall Performance on ScIntBench}
\noindent {\bf Static validity does not imply interaction readiness.} As shown in~\cref{tab:overall_evaluation}, SAGE achieves the best static physics (\textit{COL} 0.67\%, \textit{STAB} 98.38\%), yet its \textit{ISR} (6.35\%) is lower than that of Holodeck (9.41\%), whose static scores are considerable worse. Overall, 7 of the 14 methods cannot execute any prescribed interaction, and even the best completes only 36.99\%. As illustrated in~\cref{fig:qualitative_interaction} (more in Supplementary~\cref{sec:additional_results}), visually plausible scenes lead to different outcomes: only Claude Code allows opening of the refrigerator to the required angle, and only SceneSmith and Claude Code successfully generate a table that moves.

\vspace{1mm}
\noindent {\bf Interaction readiness aligns with semantic functionality.} Among the 7 methods with executable interactions, the \textit{ICaP} ranking matches \textit{Func} except that SAGE and PAT3D swap. \textit{ISR} follows a similar order except for SAGE. \textit{ICaP} also shows a shifting bottleneck: the other seven mostly fail at target existence or interactablity (\textit{ICaP} $< 2$), whereas coding agents initiate and advance interactions (\textit{ICaP} $> 6.5$) but complete only about one third. This indicates that their remaining failures lie in physical configuration.

\vspace{1mm}
\noindent {\bf Interaction success does not guarantee physical integrity.} Claude Code achieves the highest \textit{ISR} but a relatively high \textit{PIV} (15.03\%), whereas SAGE obtains the lowest \textit{PIV} (0.97\%) despite completing few tasks. Similarly, Codex achieves the highest \textit{DP} despite a lower \textit{ISR}. Thus, task completion and physical behavior capture complementary aspects of interaction readiness.

\vspace{1mm}
\noindent {\bf General-purpose coding agents lead in interaction readiness.} Claude Code and Codex rank first and second in both semantic (\textit{Func}, \textit{Lay}) and interaction (\textit{ISR}, \textit{ICaP}) metrics, and achieve the highest \textit{EXE}. But they obtain lower \textit{Comp} than SAGE and SceneSmith, which generate far richer scenes (48 and 41 vs.\ 18 and 24 objects per scene; Supplementary~\cref{tab:complete_evaluation}). We hypothesize that retrieval-based pipelines favor scene richness but inherit the limited simulation-ready coverage of their asset libraries. In contrast, coding agents can program articulations and physical properties at the cost of richness. Coding agents are also faster, requiring 25.7 and 31.4 minutes per scene (\cref{tab:computational_cost}), substantially less than SAGE and SceneSmith.

\begin{table}[t]
  \caption{Computational cost per scene. $^\dagger$API cost estimated from token usage at official pricing.}
  \label{tab:computational_cost}
  \centering
  \resizebox{\linewidth}{!}{
  \begin{tabular}{@{}lrrrrrr@{}}
    \toprule
    \multirow{2}{*}{Method}
    & \multicolumn{3}{c}{Runtime (min)}
    & \multicolumn{3}{c}{API Cost (USD)} \\
    \cmidrule(lr){2-4}
    \cmidrule(l){5-7}
    & Mean & Min & Max & Mean & Min & Max \\
    \midrule
    ATISS & 0.27 & 0.15 & 0.72 & 0 & 0 & 0 \\
    DiffuScene & 1.08 & 0.51 & 1.29 & 0 & 0 & 0 \\
    InstructScene & 0.26 & 0.03 & 0.66 & 0 & 0 & 0 \\
    PhyScene & 2.09 & 1.27 & 3.08 & 0 & 0 & 0 \\
    \midrule
    LayoutGPT & 0.41 & 0.10 & 1.98 & 0.13 & 0.08 & 0.55 \\
    LayoutVLM & 1.41 & 0.66 & 3.49 & 0.15 & 0.10 & 0.47 \\
    Holodeck & 3.61 & 2.36 & 4.82 & 0.05 & 0.03 & 0.08 \\
    HSM & 9.04 & 3.53 & 20.98 & 0.64 & 0.22 & 1.90 \\
    PAT3D & 46.73 & 14.21 & 88.03 & 1.12 & 0.28 & 2.83 \\
    \midrule
    SceneWeaver & 114.09 & 5.61 & 209.63 & 0.64 & 0.09 & 2.01 \\
    SAGE & 113.99 & 36.68 & 370.00 & 0.40 & 0.17 & 0.99 \\
    SceneSmith & 148.33 & 56.05 & 455.42 & 11.42 & 2.97 & 24.77 \\
    \midrule
    Codex & 25.66 & 15.45 & 36.66 
    & 6.54$^\dagger$ & 4.32$^\dagger$ & 10.05$^\dagger$ \\
    Claude Code & 31.39 & 16.28 & 59.93 
    & 4.65$^\dagger$ & 3.06$^\dagger$ & 6.91$^\dagger$ \\
    \bottomrule
  \end{tabular}
  }
\end{table}

\subsection{Benchmark Validation}
\noindent {\bf Validation on Interactive Reference Scenes.} We further evaluate our active interaction metrics on reference scenes from ReplicaCAD~\cite{szot2021habitat}, iGibson~\cite{li22b} (both reconstructed from real-world scans), and BEHAVIOR-1K~\cite{li2023behavior} (curated interactive simulations). We report 19 matched task assignments on which the reference scenes all succeed (9 in BEHAVIOR-1K, 3 in ReplicaCAD, and 7 in iGibson), whereas each evaluated generation method achieves at most 10 confirmed successes (about half) on the same assignments. These results show that our evaluation identifies interaction capability gaps between generated and reference scenes, highlighting limitations that can inform simulation-aware scene refinement. More details of validation on reference scenes are provided in Supplementary~\cref{sec:reference_scenes}.

\vspace{1 mm}
\noindent {\bf Human Evaluation.}
To validate the quality of LLM-generated interaction tasks, we conduct a human evaluation on 100 tasks sampled across room types and interaction categories. Four annotators independently assess task relevance, task physical feasibility, and success criterion validity. They approve 100\% of tasks as relevant, 99.7\% as physically feasible, and 95.3\% as having valid success criteria, with high inter-annotator agreement. Results show that the generated tasks are relevant, feasible, and valid.

We further validate the alignment of our MLLM-based active evaluation metrics (ICaP and DP) with human judgments using 100 interaction cases. They also sampled from different methods, action types, and execution outcomes. Four annotators independently assess ICaP using the same six-level rubric as the MLLM evaluator and DP based on interaction videos. We measure the correlation between MLLM scores and mean human ratings using Spearman's $\rho$ and Kendall's $\tau$. MLLM scores correlate strongly with human ratings ($\rho = 0.818$ for ICaP and $0.683$ for DP). Detailed protocols and results for both human studies are provided in Supplementary~\cref{sec:human_eval}.

\subsection{Interaction Failure Analysis}
\label{sec:failure}

\noindent {\bf Missing interaction capabilities.} Among 8,176 scene--interaction pairs, 5,890 (72.0\%) cannot be executed due to missing target objects or lack native physical prerequisites, \eg, SAGE's fridge lacks the articulation required to open its door (\cref{fig:qualitative_interaction}). This bottleneck shifts as methods improve (Supplementary~\cref{fig:nsa_breakdown}): trained generators mostly omit targets, specialized agentic pipelines mostly lacks physical properties, and coding agents mostly fail on accessibility.

\vspace{1mm}
\noindent {\bf Incomplete execution.} Executable interactions may still fail to achieve their goals. Claude Code executes 474 of 584 tasks and typically initiates the interaction (\textit{ICaP} of 7.30), yet achieves an \textit{ISR} of only 36.99\%. In the refrigerator-opening example, SceneSmith and Codex open the door by only $0.47^\circ$ and $0.11^\circ$, far below the required $60^\circ$. Such gaps show the need to distinguish interaction availability from actual task completion.

\vspace{1mm}
\noindent {\bf Post-interaction physical violations.} Executed interactions can also expose interpenetration, support loss, and unintended object displacement. As illustrated in~\cref{fig:overview}(c), two non-target books on a table fall onto the floor after a chair-pushing interaction. These failures highlight the need for scene generation methods to incorporate physical capabilities and maintain physical reliability during interaction.

\section{Interaction-aware Scene Generation}
\begin{figure*}[t]
    \centering
    \includegraphics[width=0.8\linewidth]{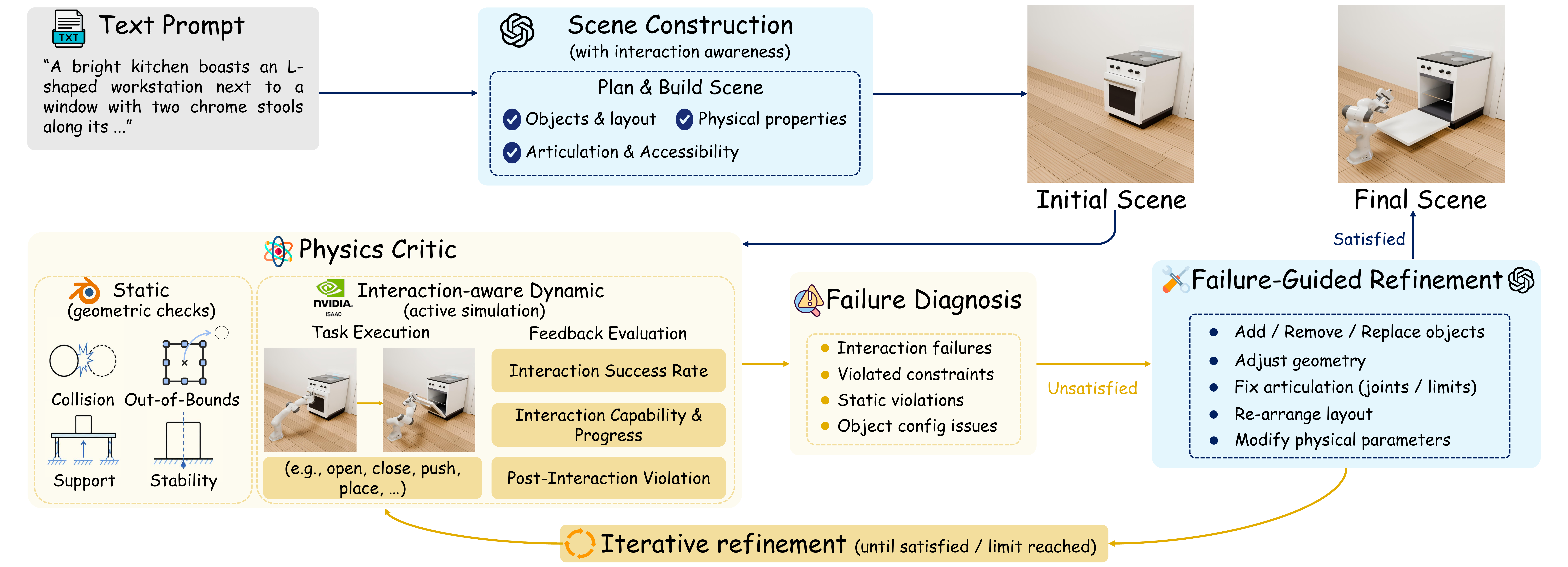}
    \caption{\textbf{Overview of our interaction-aware method.} A general-purpose coding agent builds scenes, and a physics critic with static and interactive checks diagnoses failures to guide iterative refinement.}
    \label{fig:ours}
\end{figure*}

As shown in \cref{sec:failure}, the major bottlenecks are missing interaction targets, incomplete physics configurations, and physical violations during interaction. To address them, we propose a preliminary interaction-aware scene generation method (\cref{fig:ours}).

Following agentic pipelines such as SAGE~\cite{Xia_2026_CVPR} and SceneSmith~\cite{pfaff2026scenesmith}, we replace their LLM agents and asset retrieval with a general-purpose coding agent (Codex) that constructs and iteratively refines scenes. Given a text prompt, the agent first builds an initial scene that accounts for interaction requirements, including physical properties, articulation, and accessibility. Beyond the static checks used in prior methods, our physics critic introduces an interaction-aware dynamic critic, which executes interaction tasks in simulation (planned by the agent from the text prompt, rather than the benchmark tasks) to evaluate success rate, physical behavior during execution, and scene integrity afterward. Unlike SceneCode~\cite{wang2026scenecode}, which also programs articulated objects but refines them using code errors and visual critique, and EmbodiedSmith~\cite{qin2026embodiedsmith}, which revises task plans for robotic feasibility, our critic feeds back interaction outcomes to improve the interaction readiness of the scene itself. Since objects and their physics configurations are programmed, the agent can modify geometry, articulation, physical properties, and layout. The loop repeats for up to three rounds or until all criteria are satisfied.

\begin{table*}[t]
\vspace{-3mm}
\centering
\caption{Comparison with strong baselines and ablations of our method. \#Obj: average number of objects per scene. Results for all methods are in Supplementary~\cref{tab:complete_evaluation}. \textbf{Bold} and \underline{underline} indicate the best and second-best results within each group.}
\label{tab:refinement_ablation}
\setlength{\tabcolsep}{3pt}
\renewcommand{\arraystretch}{1.15}
\resizebox{\linewidth}{!}{%
\begin{tabular}{lc*{8}{c}>{\color{gray}}c*{4}{c}}
\toprule
Method & \#Obj &
Real $\uparrow$ & Func $\uparrow$ & Lay $\uparrow$ & Comp $\uparrow$ &
COL $\downarrow$ & OB $\downarrow$ & SUP $\uparrow$ & STAB $\uparrow$ &
\multicolumn{1}{c}{EXE $\uparrow$} & ISR $\uparrow$ & ICaP $\uparrow$ & PIV $\downarrow$ & DP $\uparrow$ \\
\midrule
\multicolumn{15}{l}{
    \textit{Comparison with strong baselines}} \\

SAGE
    & 48
    & \mstd{4.88}{1.33}
    & \mstd{4.46}{1.63}
    & \mstd{3.02}{1.24}
    & \underline{\mstd{4.74}{1.98}}
    & \underline{\mstd{0.67}{1.40}}
    & \textbf{\mstd{0.00}{0.00}}
    & \mstd{76.63}{12.46}
    & \underline{\mstd{98.38}{3.31}}
    & 315/584
    & \mstd{6.35}{8.92}
    & \mstd{4.67}{1.57}
    & \textbf{\mstd{0.97}{1.30}}
    & \mstd{6.31}{0.91}\\

SceneSmith
    & 41
    & \underline{\mstd{6.80}{1.34}}
    & \mstd{6.78}{1.88}
    & \mstd{5.80}{1.55}
    & \textbf{\mstd{5.26}{1.65}}
    & \mstd{8.35}{8.01}
    & \mstd{4.67}{4.58}
    & \underline{\mstd{94.44}{7.07}}
    & \mstd{72.32}{26.14}
    & 419/584
    & \mstd{21.64}{18.38}
    & \mstd{5.81}{1.45}
    & \mstd{6.29}{6.61}
    & \mstd{6.72}{0.73}\\

Codex
    & 18
    & \mstd{6.74}{0.80}
    & \underline{\mstd{8.36}{1.06}}
    & \underline{\mstd{7.40}{1.14}}
    & \mstd{4.46}{1.30}
    & \mstd{3.48}{10.29}
    & \mstd{4.92}{9.51}
    & \mstd{85.42}{19.05}
    & \mstd{97.75}{6.85}
    & 470/584
    & \mstd{31.93}{14.74}
    & \mstd{6.58}{0.95}
    & \mstd{7.70}{9.71}
    & \underline{\mstd{6.77}{0.51}}\\

Claude Code
    & 24
    & \textbf{\mstd{7.08}{0.80}}
    & \textbf{\mstd{8.62}{1.09}}
    & \textbf{\mstd{7.62}{0.81}}
    & \mstd{4.70}{1.28}
    & \mstd{1.06}{3.68}
    & \mstd{0.40}{2.83}
    & \mstd{94.05}{8.88}
    & \mstd{92.30}{21.15}
    & 474/584
    & \underline{\mstd{36.99}{20.21}}
    & \underline{\mstd{7.30}{1.07}}
    & \mstd{15.03}{14.73}
    & \mstd{6.63}{1.16}\\
    
\textbf{Ours (Full)}
    & 16
    & \mstd{6.43}{0.99}
    & \mstd{7.80}{1.45}
    & \mstd{7.17}{1.15}
    & \mstd{3.90}{0.99}
    & \textbf{\mstd{0.00}{0.00}}
    & \underline{\mstd{0.35}{1.92}}
    & \textbf{\mstd{97.26}{7.00}}
    & \textbf{\mstd{100.00}{0.00}}
    & 492/584
    & \textbf{\mstd{43.03}{16.76}}
    & \textbf{\mstd{7.52}{1.21}}
    & \underline{\mstd{3.34}{5.25}}
    & \textbf{\mstd{6.83}{0.64}}\\
    
\midrule
\multicolumn{15}{l}{
    \textit{Controlled ablations}} \\
w/o Static Critic
    & 16
    & \mstd{6.33}{1.05}
    & \underline{\mstd{7.77}{1.45}}
    & \underline{\mstd{7.20}{0.85}}
    & \underline{\mstd{3.97}{1.30}}
    & \mstd{0.95}{5.22}
    & \underline{\mstd{0.18}{0.96}}
    & \mstd{96.79}{7.19}
    & \mstd{97.44}{3.04}
    & 483/584
    & \mstd{37.46}{21.46}
    & \mstd{7.17}{1.39}
    & \underline{\mstd{5.77}{6.15}}
    & \mstd{6.18}{1.28}\\

w/o Dynamic Critic
    & 16
    & \mstd{6.27}{0.92}
    & \mstd{7.67}{1.56}
    & \mstd{6.87}{1.11}
    & \mstd{3.87}{0.90}
    & \textbf{\mstd{0.00}{0.00}}
    & \mstd{0.32}{1.92}
    & \mstd{97.01}{5.08}
    & \mstd{99.43}{3.09}
    & 476/584
    & \mstd{35.73}{21.08}
    & \underline{\mstd{7.26}{1.34}}
    & \mstd{7.57}{9.73}
    & \underline{\mstd{6.70}{0.63}}\\

w/o Diagnosis
    & 17
    & \underline{\mstd{6.40}{0.71}}
    & \mstd{7.73}{1.55}
    & \textbf{\mstd{7.33}{0.84}}
    & \textbf{\mstd{4.00}{1.05}}
    & \textbf{\mstd{0.00}{0.00}}
    & \textbf{\mstd{0.00}{0.00}}
    & \underline{\mstd{97.21}{5.93}}
    & \textbf{\mstd{100.00}{0.00}}
    & 480/584
    & \underline{\mstd{38.54}{18.31}}
    & \mstd{7.20}{1.31}
    & \mstd{6.15}{8.37}
    & \mstd{6.58}{0.43}\\

\textbf{Full}
    & 16
    & \textbf{\mstd{6.43}{0.99}}
    & \textbf{\mstd{7.80}{1.45}}
    & \mstd{7.17}{1.15}
    & \mstd{3.90}{0.99}
    & \textbf{\mstd{0.00}{0.00}}
    & \mstd{0.35}{1.92}
    & \textbf{\mstd{97.26}{7.00}}
    & \textbf{\mstd{100.00}{0.00}}
    & 492/584
    & \textbf{\mstd{43.03}{16.76}}
    & \textbf{\mstd{7.52}{1.21}}
    & \textbf{\mstd{3.34}{5.25}}
    & \textbf{\mstd{6.83}{0.64}}\\
\bottomrule
\end{tabular}%
}
\vspace{-3mm}
\end{table*}

\subsection{Comparison with Baselines}
We compare our method with the original Codex under the same generation settings and strong baselines from~\cref{tab:overall_evaluation}. \Cref{tab:refinement_ablation} shows our method eliminates collisions and instability, raises \textit{ISR} from 31.93\% to 43.03\%, outperforming other methods, and reduces \textit{PIV} from 7.70\% to 3.34\%. This comes at a modest cost in semantic scores (\eg, \textit{Comp} from 4.46 to 3.90), but \textit{EXE} increases from 470 to 492, indicating that interaction targets are preserved. Runtime increases over Codex but remains below SceneSmith (\cref{tab:ablation_efficiency}).

\begin{table}[t]
\centering
\caption{Computational cost per scene: runtime and number of repair candidates. Full results are in Supplementary~\cref{tab:overall_efficiency}.}
\label{tab:ablation_efficiency}
\footnotesize
\setlength{\tabcolsep}{6pt}
\begin{tabular*}{\linewidth}
{@{\extracolsep{\fill}}lcc@{}}
\toprule
Method &
Runtime (min) &
Repair Candidates \\
\midrule
SceneSmith & 148.33 & 18.9\\
Codex & 25.66 & 4.6\\
\midrule
w/o Static Critic  & 115.29 & 10.7 \\
w/o Dynamic Critic & 19.46 & 2.8 \\
w/o Diagnosis      & 128.22 & 14.7 \\
\textbf{Full}     & 119.35 & 11.8 \\
\bottomrule
\end{tabular*}
\vspace{-2mm}
\end{table}

\subsection{Ablation Study}
We ablate each component of our method (\cref{tab:refinement_ablation}): the static critic, the interaction-aware dynamic critic, and failure diagnosis. Removing any component degrades all active interaction metrics, while semantic scores remain similar across variants. Without the static critic, collisions and instability reappear. Although removing the dynamic critic saves most of the runtime (\cref{tab:ablation_efficiency}), it causes the largest drop in \textit{ISR} and the highest \textit{PIV}. Without diagnosis, the agent explores more repair candidates and takes longer time, but achieves lower \textit{ISR}. This shows that explicit failure analysis makes repair both more effective and more efficient.

\section{Conclusion}
\label{sec:conclu}
In this paper, we propose ScIntBench, a systematic benchmark that extends the evaluation of 3D indoor scenes beyond semantic quality and static physical validity to interaction readiness through four new metrics. Our evaluation of 14 representative methods reveals that scenes with strong semantic and static performance do not necessarily support interaction, with many lacking the required capabilities. We also find that general-purpose coding agents emerge as competitive alternatives to specialized scene generation pipelines, although substantial gaps in interaction success and physical reliability remain. Building on these findings, we further present an interaction-aware method that uses a coding agent with a simulation-based physics critic, which outperforms the strongest baselines. We hope ScIntBench provides a standardized basis for assessing and advancing interaction readiness in 3D scene generation.

\vspace{1mm}
\noindent {\bf Limitations and future work.} Currently, ScIntBench is limited to rigid-body and articulated interactions in indoor scenes. Extending it to deformable objects, fluids, and outdoor scenes is a promising direction as scene generation methods and physics simulators mature.

\section{Acknowledgments}
\label{sec:ack}
This research is supported by the Australian Research Council (ARC) under the Discovery Project scheme (Project ID: DP240101926).

{
    \small
    \bibliographystyle{ieeenat_fullname}
    \bibliography{main}
}

\clearpage
\maketitlesupplementary
This supplementary material provides additional information, including implementation details of the evaluated baselines (\cref{sec:implementation}); construction details of ScIntBench (\cref{sec:construction}); detailed definitions, scoring criteria, and aggregation procedures for all evaluation metrics (\cref{sec:metrics_detail}); scene conversion and its verification, simulation environment, configurations, and interaction execution protocols (\cref{sec:simulation_detail}); additional results in ScIntBench (\cref{sec:additional_results}); and validation of the proposed benchmark (\cref{sec:benchmark_validation}). 
\appendix

\addtocontents{toc}{\protect\setcounter{tocdepth}{2}}
\begingroup
\setcounter{tocdepth}{-2}
\tableofcontents
\endgroup
\setcounter{tocdepth}{2}

\section{Baseline Implementation Details}
\label{sec:implementation}
In this section, we provide implementation details for the scene generation methods evaluated in the main paper. \Cref{tab:baseline_config} summarizes their training strategies, backbone models, and the retrieved asset libraries, highlighting the differences between task-specific trained generators and methods primarily relying on pretrained foundation models. All external language or vision models are accessed via API usage. In addition, the computational environments differ across methods. SAGE~\cite{Xia_2026_CVPR} and SceneSmith~\cite{pfaff2026scenesmith} are executed using two NVIDIA L40S GPUs, while the remaining executed methods primarily use one NVIDIA RTX 4090 GPU.

For general-purpose coding agents, we also document scene generation prompts, permitted tools, output specifications, and execution settings. The complete prompt templates for Codex~\cite{openai2026codex} and Claude Code~\cite{anthropic2026claudecode} are provided below.

\begin{promptbox}{Scene Generation Prompt}
\scriptsize
\textbf{Task}\quad
You are given a natural-language description of an indoor room. Autonomously construct a complete 3D scene that satisfies the description. The scene should be semantically consistent, spatially plausible, and suitable for downstream physics-based embodied interaction. You may autonomously use the available development tools, Blender, Python, and provided assets as needed. You may write and execute scripts, inspect intermediate results, and iteratively debug the scene without human assistance. Do not request human assistance during generation.

\textbf{Criteria}\quad The final scene must:
\begin{itemize}
    \item contain the major objects and furniture implied by the room description
    \item use realistic object scales, placements, and orientations
    \item avoid unintended object intersections
    \item maintain physically plausible object placement and support relationships
    \item preserve major interactive objects as separate scene objects
    \item include collision/physics properties sufficient for use in a physics simulator
    \item be inspectable and editable in Blender, and be exportable to and loadable in NVIDIA Isaac Sim
\end{itemize}

\textbf{Input Format} \quad
A natural-language description of an indoor room:
\texttt{\{room\_prompt\}}.

\textbf{Output Format} \quad
Produce the following final artifacts:
\begin{enumerate}
    \item \texttt{scene.blend}: an inspectable and editable Blender scene.
    \item An Isaac-Sim-compatible USD scene
    (\texttt{scene.usd} or equivalent).
    \item All scripts and local dependencies required to reproduce
    the generated scene.
    \item \texttt{scene\_metadata.json}.
\end{enumerate}

\end{promptbox}

\begin{table*}[t]
\centering
\caption{\textbf{Configuration of the evaluated methods.} \emph{3D-FRONT} denotes task-specific scene generation models trained on the 3D-FRONT~\cite{fu20213d} dataset, while \emph{Pretrained} denotes approaches primarily relying on pretrained foundation models. Asset sources include retrieval libraries such as 3D-FUTURE~\cite{fu20213d}, GAPartNet~\cite{geng2023gapartnet}, Objaverse~\cite{deitke2023objaverse}, HSSD~\cite{khanna2024habitat}, Infinigen~\cite{raistrick2024infinigen}, MetaScenes~\cite{yu2025metascenes}, and ArtVIP~\cite{jin2026artvip}, as well as assets generated by Hunyuan3D-2~\cite{zhao2025hunyuan3d}, TRELLIS~\cite{xiang2025structured}, and SAM3D~\cite{chen2026sam}.}
\label{tab:baseline_config}
\small
\setlength{\tabcolsep}{9pt}
\renewcommand{\arraystretch}{1.08}
\resizebox{\textwidth}{!}{%
\begin{tabular}{lllll}
\toprule
\textbf{Method} &
\textbf{Category} &
\textbf{Training} &
\textbf{Backbone} &
\textbf{Asset Source}\\
\midrule
ATISS~\cite{paschalidou2021atiss}
    & Task-specific & 3D-FRONT & -- & 3D-FUTURE\\
DiffuScene~\cite{tang2024diffuscene}
    & Task-specific & 3D-FRONT & -- & 3D-FUTURE\\
InstructScene~\cite{lin2024instructscene}
    & Task-specific & 3D-FRONT & -- & 3D-FUTURE\\
PhyScene~\cite{yang2024physcene}
    & Task-specific & 3D-FRONT & -- & 3D-FUTURE, GAPartNet\\
\midrule
LayoutGPT~\cite{feng2023layoutgpt}
    & LLM/VLM-based & Pretrained & GPT-4
    & 3D-FUTURE \\
LayoutVLM~\cite{sun2025layoutvlm}
    & LLM/VLM-based & Pretrained & GPT-4o
    & Objaverse \\
Holodeck~\cite{yang2024holodeck}
    & LLM/VLM-based & Pretrained & GPT-4
    & Objaverse \\
HSM~\cite{pun2026hsm}
    & LLM/VLM-based & Pretrained & GPT-4o
    & HSSD \\
PAT3D~\cite{lin2026pat3d}
    & LLM/VLM-based & Pretrained & GPT-5.4
    & Generated (Hunyuan3D-2) \\
\midrule
SceneWeaver~\cite{yang2026sceneweaver}
    & Specialized agentic & Pretrained & GPT-4.1 
    & 3D-FUTURE, Objaverse, Infinigen, MetaScenes\\
SAGE~\cite{Xia_2026_CVPR}
    & Specialized agentic & Pretrained
    & GPT-oss-120b, Qwen3-VL-30B-A3B-Instruct 
    & Generated (TRELLIS)\\
SceneSmith~\cite{pfaff2026scenesmith}
    & Specialized agentic & Pretrained & GPT-5.2 
    & Generated (SAM3D), ArtVIP\\
\midrule
Codex~\cite{openai2026codex}
    & Coding agent & Pretrained & GPT-6-Astra 
    & Procedural (Blender)\\
Claude Code~\cite{anthropic2026claudecode}
    & Coding agent & Pretrained & Opus-5.5 
    & Procedural (Blender)\\
\bottomrule
\end{tabular}%
}
\end{table*}

\section{Benchmark Construction Details}
\label{sec:construction}
\subsection{Scene Prompts and Room Types}
\begin{figure}[htbp!]
  \centering
   \includegraphics[width=0.8\linewidth]{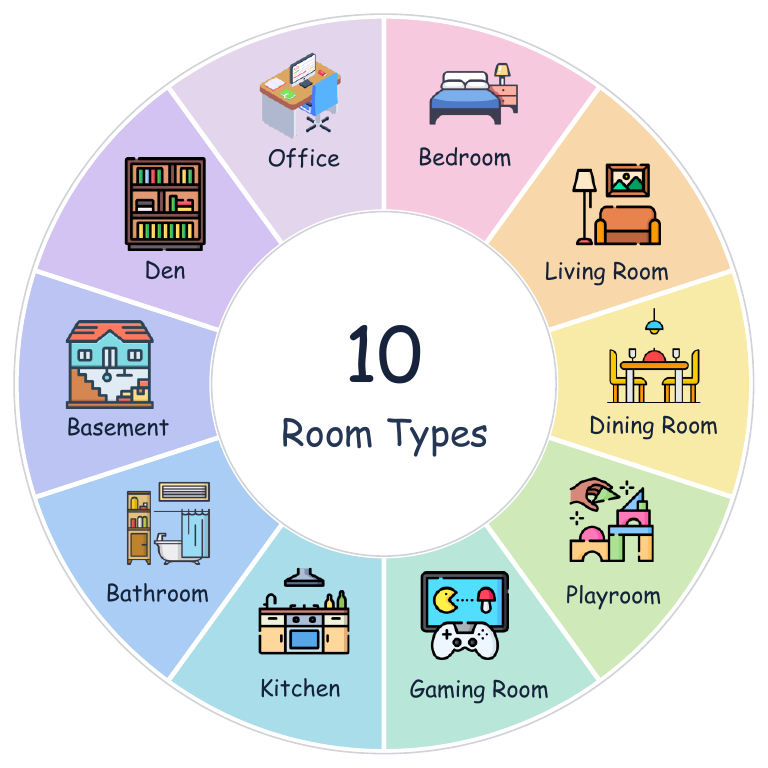}
   \caption{\textbf{Overview of the 10 room types included in ScIntBench.}}
   \label{fig:roomtype}
\end{figure}

We select 50 indoor scene descriptions from SceneEval~\cite{tam2026sceneeval}, covering 10 different room types, as shown in Fig.~\ref{fig:roomtype}. The resulting benchmark comprises 700 generated scenes.

\subsection{Interaction Task Construction}
To evaluate interaction readiness independently of scene generation methods, we construct a shared set of interaction tasks from each original text prompt. We use GPT-5.6 in a two-stage procedure. First, it identifies applicable action categories from a predefined action library and enumerates all entities explicitly mentioned in the scene description. Second, it examines each action-entity pair and determines whether the interaction is semantically applicable.

The current action library includes 7 categories: \texttt{open\_hinged}, \texttt{open\_drawer}, \texttt{close}, \texttt{push\_rigid}, \texttt{pick\_and\_place}, \texttt{stack}, \texttt{place\_push\_probe}. These interaction actions corresponding to opening hinged parts, opening drawers along their prismatic joint axes, closing hinged or sliding parts, pushing rigid objects, picking up objects and placing them on surfaces, placing one object on top of another, and surface interactions based on standardized probe. This library is used to guide the LLM in selecting appropriate actions, avoiding it from consistently favoring a single action type.

The prompt template used for interaction task generation is illustrated as below:

\begin{promptbox}{Task Generation Prompt}
\scriptsize
\textbf{Stage 1: Action and Entity Identification.}

\textbf{Task} \quad
Choose all meaningful action FAMILIES from the supplied executable library, and inventory every explicit physical entity in the room prompt (including landmarks, mounted objects and architecture). Merge repeated references to one entity. Do not invent contents, handles or implicit components. Quote a contiguous phrase for each entity. Use compatible canonical categories from the library where possible, but keep inapplicable entities with their own category. Count is the number explicitly specified (e.g. three chairs $\rightarrow$ 3), otherwise 1; unspecified plurals represent one category slot, not an invented count. Double bed and two-seater sofa are each one entity. Return the supplied JSON schema.
\textbf{Input Format}
The input will be \{room\_prompt\} and \{library\}.

\textbf{Output Format}
Return the results: \{selected\_family\} and \{entity\_inventory\}.

\medskip
\textbf{Stage 2: Action--Entity Applicability.}

\textbf{Task} \quad
Classify EVERY selected action family x inventoried entity pair, including both applicable and inapplicable pairs. Include all applicable targets, no top-K limit. Judge semantic applicability using the prompt and library, never predicted method capabilities. A missing native mechanism is checked later per method and MUST NOT prune a meaningful intended action. Push-rigid is horizontal translation of floor-supported freestanding objects and must end floor-supported; do not detach mounted fixtures, push shelf contents off supports, or treat a fixed built-in island / vanity as freestanding. Hinged opening applies only to normally hinged enclosures / doors, not generic opening of drawer-only storage; identify uncertainty in the reason. Surface probing requires a meaningful accessible horizontal support surface for the standardized small rigid probe, not an invented platform on a curved object or an opening filled by a sink. Every pair must appear exactly once.

\textbf{Input Format}
The input will be \{room\_prompt\}, \{library\}, and \{entity\_inventory\}.

\textbf{Output Format}
Return the results in the following JSON format:
\begin{lstlisting}[style=promptcode]
{"pairs": [
    {
      "action family": "...",
      "object": "...",
      "applicable": true / false,
      "reason": "..."
    },...]
}
\end{lstlisting}
\end{promptbox}

\subsection{Benchmark Example}
\begin{table*}[ht]
\centering
\caption{Interaction tasks constructed for the example scene (Dining Room). The action type, target object, and main success criteria are shown. For pick-and-place and stacking tasks, arrows indicate the source object and destination support.}
\label{tab:benchmark_example}
\small
\setlength{\tabcolsep}{5pt}
\renewcommand{\arraystretch}{1.08}
\begin{tabular}{llll}
\toprule
\textbf{Action} &
\textbf{Action Family} &
\textbf{Target Object / Destination} &
\textbf{Success Criteria} \\
\midrule

Open buffet door
& \multirow{3}{*}{\texttt{open\_hinged}}
& Buffet
& \multirow{3}{*}{\shortstack[l]{
    Opening $\geq 60^\circ$,\\
    held for $\geq 1$ s}} \\
Open room door
& & Room door & \\
Open sideboard door
& & Sideboard & \\

\midrule

Push dining table
& \multirow{4}{*}{\texttt{push\_rigid}}
& Dining table
& \multirow{4}{*}{\shortstack[l]{
    Displacement $0.13$--$0.17$ m\\
    along the prescribed direction,\\
    maintained for $\geq 1$ s}} \\
Push chair
& & Chair & \\
Push buffet
& & Buffet & \\
Push sideboard
& & Sideboard & \\

\midrule

\multirow{3}{*}{Place and push probe}
& \multirow{3}{*}{\texttt{place\_push\_probe}}
& Dining-table surface
& \multirow{3}{*}{\shortstack[l]{
    Supported sliding $\geq 0.02$ m\\
    for $\geq 0.2$ s}} \\
& & Buffet top & \\
& & Sideboard top & \\

\midrule

\multirow{6}{*}{Pick and place}
& \multirow{6}{*}{\texttt{pick\_and\_place}}
& Candelabra $\rightarrow$ buffet
& \multirow{6}{*}{\shortstack[l]{
    Lift $\geq 0.05$ m;\\
    horizontal relocation $\geq 0.10$ m;\\
    release fully inside the\\
    designated destination region;\\
    stable support for $\geq 1$ s}} \\
& & Fruit bowl $\rightarrow$ dining table & \\
& & Plate $\rightarrow$ dining table & \\
& & Wine glass $\rightarrow$ dining table & \\
& & Decanter $\rightarrow$ sideboard & \\
& & Cognac glass $\rightarrow$ sideboard & \\

\midrule

\multirow{2}{*}{Stack plates}
& \multirow{2}{*}{\texttt{stack}}
& \multirow{2}{*}{Plate 2 $\rightarrow$ plate 1}
& Lift $\geq 0.05$ m; center within receiver support; \\
& & & stable for $\geq 1$ s after release \\

\midrule

Close buffet door
& \multirow{3}{*}{\texttt{close}}
& Buffet
& \multirow{3}{*}{\shortstack[l]{
    Closed-reference error $\leq 5^\circ$,\\
    maintained for $\geq 1$ s\\
    after release}} \\
Close room door
& & Room door & \\
Close sideboard door
& & Sideboard & \\

\bottomrule
\end{tabular}
\end{table*}

We present an example of ScIntBench as below:
\paragraph{Room Type:}\textit{Dining Room.}

\paragraph{Text Prompt:}\textit{In a formal dining room, a long mahogany rectangular table rests at the center with eight navy velvet chairs---four on each long side. A crystal chandelier hangs overhead from the ceiling. Along the back wall, a buffet displays two silver candelabras, topped by a large gold-framed painting centered above. Sunlight filters through a tall window opposite the door. On the table, a porcelain fruit bowl sits beside a floral centerpiece in the middle. Two white porcelain
plates are set with linen napkins to their right, and two wine glasses line nearby. A sideboard beside the door holds a crystal decanter and two cognac glasses.}

\paragraph{Interaction Tasks:}\cref{tab:benchmark_example} summarizes the interaction tasks constructed for this scene across six action families, including their action families, target objects, and main success criteria. For pick-and-place and stack tasks, each object is relocated to a designated region of its specified support surface.

\section{Metrics Details}
\label{sec:metrics_detail}
We evaluate generated scenes in three complementary dimensions: semantic quality, static physical plausibility, and active interaction behavior. This section provides details of implementation and scoring for the metrics introduced in the main paper. All evaluations operate on the generated scene content without adding missing physical capabilities to improve the evaluated results.

\subsection{Semantic Quality Evaluation}
\label{sec}
Following SceneWeaver~\cite{yang2026sceneweaver}, semantic quality is assessed using four criteria: realism, functionality, layout, and completeness. Each criterion is evaluated on a 0--10 scale using a MLLM-based judge.

For each generated scene, we render five standardized views using Blender: one top-down view and four corner views. Each image has a resolution of 1024 $\times$ 1024. All five views are provided, together with the original scene description.

The four dimensions are defined as follows. \textit{Realism} measures the visual plausibility of the generated environment. \textit{Functionality} assesses whether the furnishings and spatial arrangement are suitable for the intended room activities. \textit{Layout} measures the coherence of object placement and spatial organization. \textit{Completeness} (denoted as "completion" in the prompt) evaluates whether the scene contains sufficient objects and details to satisfy the text description and form a complete indoor environment. The complete scoring instructions for the MLLM are provided below:
\begin{promptbox}{Semantic Quality Scoring Prompt}
\scriptsize
\textbf{Task}\quad
You are given a top-down and four corner-view room render images. Your task is to evaluate how well they align with the user's preferences across the four criteria listed below. For each criterion, assign a score from 0 to 10, and provide a brief justification for your rating. Scoring must be strict. If any critical issue is found (such as missing key objects, obvious layout errors, or unrealistic elements), the score should be significantly lowered, even if other aspects are fine.

\textbf{Score Guidelines}
\begin{itemize}
  \item Score 10: Fully meets or exceeds expectations; no major improvements needed.
  \item Score 5: Partially meets expectations; some obvious flaws exist that limit usefulness.
  \item Score 0: Completely fails to meet expectations; the aspect is absent, wrong,
  or contradicts user needs.
\end{itemize}

\textbf{Evaluation Criteria}
\begin{enumerate}
  \item \textbf{Realism:}
  How realistic the room appears. Ignore texture, lighting, and doors.
  \begin{itemize}
    \item Good (8-10): The layout (position, rotation, and size) is believable, and common
    daily objects make the room feel lived-in. Rich of daily furniture and objects.
    \item Bad (0-3): Unusual objects or strange placements make the room unrealistic.
    \item Note: If object types or combinations defy real-world logic
    (e.g., bathtubs in bedrooms), score should be below 5.
  \end{itemize}

  \item \textbf{Functionality:}
  How well the room supports the intended activities.
  \begin{itemize}
    \item Good (8-10): Contains the necessary furniture and setup for the specified function.
    \item Bad (0-3): Missing key objects or contains mismatched furniture
    (e.g., no bed in a bedroom).
    \item Note: Even one missing critical item should lower the score below 6.
  \end{itemize}

  \item \textbf{Layout:}
  Whether the furniture is arranged logically in good pose and aligns with the user's preferences.
  \begin{itemize}
    \item Good (8-10): Each objects is in reasonable size, neatly placed, objects of the
    same category are well aligned, relationships are reasonable (e.g., chairs faces desks),
    sufficient space exists for walking, and orientations must be correct.
    \item Bad (0-3): Floating objects, crowded floor, abnormal size, objects with collision,
    incorrect orientation, or large items placed oddly (e.g., sofa not against the wall).
    Large empty space. Blocker in front of furniture.
    \item Note: If the room has layout issues that affect use, it should not score above 5.
  \end{itemize}

  \item \textbf{Completion:}
  How complete and finished the room feels.
  \begin{itemize}
    \item Good (8-10): All necessary large and small items are present. Has rich details.
    Each shelf has multiple objects inside. Each supporter (e.g., table, desk, and shelf)
    has small objects on it. Empty area is less than 50\%. The room feels done.
    \item Bad (0-3): Room is sparse or empty, lacks decor or key elements.
    \item Note: If more than 50\% of the room is blank or lack detail, score under 5.
  \end{itemize}
\end{enumerate}

\textbf{Input Format}
The input will be a \{scene\_description\} and \{rendered\_images $\mathcal{I}_t$\}.

\textbf{Output Format}
Return the results in the following JSON format, the comment should be short:
\begin{lstlisting}[style=promptcode]
{ "realism": {
    "grade": your grade as int,
    "comment": "Your comment and suggestion."
  },
  "functionality": {...},
  "layout": {...},
  "completion": {...}}
\end{lstlisting}
\end{promptbox}

\subsection{Static Physical Evaluation}
\label{sec}
Static physical plausibility is evaluated using four metrics: collision, out-of-bounds, support correctness (following SceneEval~\cite{tam2026sceneeval}), and stability (following SAGE~\cite{Xia_2026_CVPR}).

\paragraph{Collision (COL).} We use SceneEval's~\cite{tam2026sceneeval} mesh-based collision detector, including its 5-mm contact-filtering procedure: an initially intersecting object pair is counted as colliding only if the intersection persists after the prescribed temporary separation check. This operation is performed on a detection copy and does not modify the generated scene. 

\paragraph{Out-of-Bounds (OB).} We adopt SceneEval's~\cite{tam2026sceneeval} out-of-bounds detector, which samples points on each object's mesh surface and casts downward rays toward the evaluation floor. An object is classified as out of bounds when fewer than 99\% of its sampled points intersect the floor. 

\paragraph{Support (SUP).} Following SceneEval~\cite{tam2026sceneeval}, we assess support in two stages. First, GPT-5.6 determines the expected support category of each object from multi-view visual evidence, selecting among ground, object, wall, and ceiling support. Second, we verify the corresponding geometric support relationships using an adapted ray-based contact detector.

\paragraph{Stability (STAB).} Following the stability criterion used in SAGE~\cite{Xia_2026_CVPR}, we simulate each scene for 120 physics steps and identify objects whose displacement exceeds 0.2 m or whose orientation change exceeds $8^\circ$ relative to the initial state. Our evaluation uses Isaac Sim 5.1. 

\subsection{Interaction Success Rate (ISR)}
The Interaction Success Rate measures the proportion of prescribed tasks that satisfy their predefined success conditions under the standardized interaction protocol. A task is considered successful only when its numerical success criteria are met. If the generated scene lacks the required target object or native physical prerequisites, the corresponding task is counted as unsuccessful.

For a scene with $N_p$ prescribed tasks, ISR is defined as:
\begin{equation}
\mathrm{ISR}_{m,p}
=
\frac{1}{{N}_p}
\sum_{t\in\mathcal{T}_p}
\mathbb{I}[\mathrm{Success}(m,p,t)],
\end{equation}
where $m$ denotes the generation method, $\mathcal{T}_p$ denotes the set of shared tasks for prompt $p$. The indicator function equals 1 if the task succeeds
and 0 otherwise.

\subsection{Interaction Capability Progress (ICaP)}
Interaction Capability Progress provides a more fine-grained assessment of interaction readiness than binary success. We use a six-level ordinal rubric to distinguish target availability, physical interaction capability, execution progress, and successful task completion.

\begin{promptbox}{ICaP Rubric}
\scriptsize
\textbf{Score Guidelines}
For the intended interaction action:
\begin{itemize}
  \item Score 10: Task Success. The interaction reaches the predefined success state.
  \item Score 8: Substantial Progress. The interaction produces substantial and physically correct progress toward the intended outcome, but does not reach the predefined success state.
  \item Score 6: Interaction Initiated. The intended interaction can be physically initiated and produces a valid response, but the execution fails before making substantial progress toward the predefined success state.
  \item Score 4: Physically Interactable. The interaction target contains the required physical properties or mechanisms, such as valid collision geometry, joints, or articulation, such that the intended action is physically possible.
  \item Score 2: Target Present. The required interaction target is present in the scene, but does not provide the physical structure or properties required for the intended action.
  \item Score 0: Target Absent. The required interaction target is absent from the scene.
\end{itemize}
\end{promptbox}

The MLLM judge receives task-specific evidence, including the scene representation, available native capabilities, and execution observations when applicable. Its assessment follows the same ordered rubric for all generation methods and action categories. Hence, for a scene with $N_p$ tasks, ICaP is measured as:
\begin{equation}
\mathrm{ICaP}_{m,p}
=
\frac{1}{{N}_p}
\sum_{t\in\mathcal{T}_p}
s(m,p,t),
\end{equation}
where $s(m,p,t) \in \{0,2,4,6,8,10\}$ denotes the score assigned to task $t$.

\subsection{Post-Interaction Physical Violation (PIV)}
Post-Interaction Physical Violation measures newly introduced or worsened physical inconsistencies observed after an executed interaction. We compare the initial scene state with the state obtained after execution. We consider violations including object interpenetration, out-of-bounds conditions, unintended displacements, and loss of required support relationships. For each object in the scene, we determine whether at least one disallowed violation has newly appeared or worsened. Also, an object affected by multiple violation types is counted only once.

For each executed task $t$, the violation rate is defined as:
\begin{equation}
    \mathrm{Violation}(m,p,t) = \frac{|\mathcal{V}_{m,p,t}|}
{|\mathcal{O}_{m,p}|},
\end{equation}
where $\mathcal{O}_{m,p}$ denotes the set of objects of the original scene, and $\mathcal{V}_{m,p,t}$ denotes the subset that exhibits newly introduced or worsened physical violations after task $t$.

Then, the scene-level PIV for a scene is defined as:
\begin{equation}
    \mathrm{PIV}_{m,p}
=
\frac{1}{N'_{m,p}}
\sum_{t\in\mathcal{T}'_{m,p}}
\mathrm{Violation}(m,p,t),
\end{equation}
where $\mathcal{T}'_{m,p}$ denotes the set of interaction tasks that have been validly executed, and $N'_{m,p}=|\mathcal{T}'_{m,p}|$.

Furthermore, those intended task motions are distinguished from physical violations. For example, opening a door or translating the intended target does not itself constitute a violation. However, intended motion does not exempt an object from collision, out-of-bounds, or support checks.

\subsection{Dynamic Plausibility (DP)}
Dynamic Plausibility evaluates whether the observed physical behavior of a scene is consistent with plausible interaction dynamics. For each executed interaction, we record target-focused and overview observations that cover the interaction progress and the post-interaction settling period. The MLLM evaluates the visual evidence from action render video frames using a fixed 0--10 scoring rubric.

\begin{promptbox}{Dynamic Plausibility Scoring Prompt}
\scriptsize
\textbf{Task}\quad
You are given frames from an interaction video recorded in a physics simulator, the intended action, and its execution protocol. Evaluate Dynamic Plausibility (DP): "How physically plausible are the observed scene-object behaviors throughout
the interaction?" Be a strict, evidence-grounded evaluator. High scores must be earned by visible physical behavior.

\textbf{Physical Criteria}
\begin{enumerate}
    \item \textbf{Motion and Causality.} Assess whether object motion follows contact or applied forces, with plausible acceleration, inertia, and stopping. Identify unexplained motion, discontinuities, or violations of stated constraints. Smooth motion alone does not imply correct causality.
    \item \textbf{Contact and Collision.} Examine contact interactions across frames. Identify interpenetration, clipping, pass-through, chattering, unrealistic impacts, or missing physical coupling. Occluded contacts are unverified, and visual artifacts alone do not establish collisions.
    \item \textbf{Gravity and Support.} Evaluate falling, settling, tipping, and support transitions, including non-target objects. A single significant violation matters. Physically coherent falling or damage is not inherently a violation; assess the underlying physical behavior.
    \item \textbf{Friction and Sliding.} Evaluate resting, sliding onset, sustained motion, and deceleration against applied forces. Avoid inferring exact physical parameters or inventing hidden constraints to explain unexpected motion. Acknowledge insufficient evidence.
    \item \textbf{Rigidity and Articulation.} Assess rigid-body integrity, attachments, hinge axes, and joint motion. Identify unintended detachment, deformation, pivot changes, or penetration. Respect stated rigid-body approximations.
\end{enumerate}

\textbf{Score Guidelines}
\begin{itemize}
  \item High Plausibility (8-10): Critical interaction phases are clearly observed and physically coherent, with no moderate or major violations. Minor defects or noncritical uncertainty may remain. A score of 10 requires clear positive evidence of physical plausibility.
  \item  Moderate Plausibility (4-7): The interaction remains broadly understandable but contains minor to moderate physical inconsistencies or partially observed critical phases. Scores decrease with the severity, frequency, and extent of inconsistencies. Missing evidence should not be treated as a physical violation.
  \item Low Plausibility (0-3): Clear major physical violations undermine the interaction, ranging from isolated major failures (3) to pervasive physical incoherence (0). A single major violation limits the score to 3 or below. Missing targets and task failures do not justify a score of 0.
\end{itemize}
\textbf{Input Format}
The input will be a \{action\} and \{rendered\_video\_frames\}.

\textbf{Output Format}
Return the results in the following JSON format, the comment should be short:
\begin{lstlisting}[style=promptcode]
{
  "score": you grade as int,
  "comment": "Brief justification identifying the decisive evidence and limit.",
  "violations": [],
}
\end{lstlisting}
\end{promptbox}

During the judgment, task failure does not automatically imply low dynamic plausibility: an interaction may fail to reach its target while exhibiting physically reasonable behavior. Conversely, successful task completion does not guarantee physical plausibility if visible violations occur.

\subsection{Metric Aggregation and Evaluation Coverage}
For each scene, the interaction-level scores are aggregated over the corresponding shared task set. ISR and ICaP account for the prescribed interactions, including cases in which the target object or required native capability is missing. PIV and DP are computed conditionally on valid executed interactions, and their reported averages are therefore accompanied by evaluation coverage.

Across the 50 scene descriptions and 14 generation methods, ScIntBench evaluates 8,176 prescribed method--interaction pairs. Of these, 2,286 interactions are executed with valid physical evidence, while 5,890 are classified as unsupported by the generated scene.

\section{Standardized Evaluation Protocol}
\label{sec:simulation_detail}

\subsection{Scene Conversion and Loading}  
We render scenes in Blender~\cite{blender} and conduct interaction evaluation and physical simulation in NVIDIA Isaac Sim 5.1~\cite{nvidiaisaacsim}. We use a common scene representation
and method-specific loading adapters to accommodate the heterogeneous outputs of the evaluated methods.

\paragraph{Scene conversion.}
The evaluated methods produce different output formats, including layout files, Blender scenes, and USD, SDF, or URDF assets. For each scene, we construct a unified JSON manifest that records object identities, geometry references, spatial transforms, and source provenance. Together with the associated mesh and material assets, this manifest supports scene reconstruction in Blender. For Isaac Sim, a method-specific adapter assembles a USD stage by reusing native USD layers when available, or by exporting geometry and translating the physical descriptions provided by the original assets and their official loading procedures. The adapter establishes correspondences between source objects and USD rigid bodies, colliders, and joints. It maps the physical properties provided by each method, including collision geometry, mass, inertia, friction, and joint constraints, to the corresponding USD entities.

\paragraph{Physical properties and verification.} 
To assess conversion accuracy and identify discrepancies that could affect further evaluation, we audit ten randomly selected scenes per method (\cref{tab:conversion_audit}). For each audited scene, we compare the records in the original source with the intermediate physics manifest, the composed USD declarations, and the effective PhysX values, checking geometry, numerical values, units, and object bindings. Mesh vertices deviate by at most 1.9\,$\mu$m from the source, which is at the level of floating-point precision. For all methods that provide physical properties, every traced property record matches its USD declaration, and all runtime checks of masses, material bindings, friction, restitution, and joints pass in PhysX. Properties that a method does not provide will not be supplemented, and are marked as N/A. Thus, these results confirm that the conversion preserves the geometry and physical properties produced by each method, so that differences in our results reflect the generated scenes rather than the conversion.

\subsection{Interaction Execution and Configurations}
We evaluate 7 interaction families: \texttt{open\_hinged}, \texttt{open\_drawer}, \texttt{close}, \texttt{push\_rigid}, \texttt{pick\_and\_place}, \texttt{stack}, \texttt{place\_push\_probe}. For the first six families, we use a Franka Panda manipulator available in Isaac Sim to interact with target objects through physical contact. For surface probing, we use a standardized rigid probe with prescribed physical properties and an external force schedule. 

\subsection{Execution Protocol}
\label{sec:execution}
\begin{figure}[t]
  \centering
   \includegraphics[width=\linewidth]{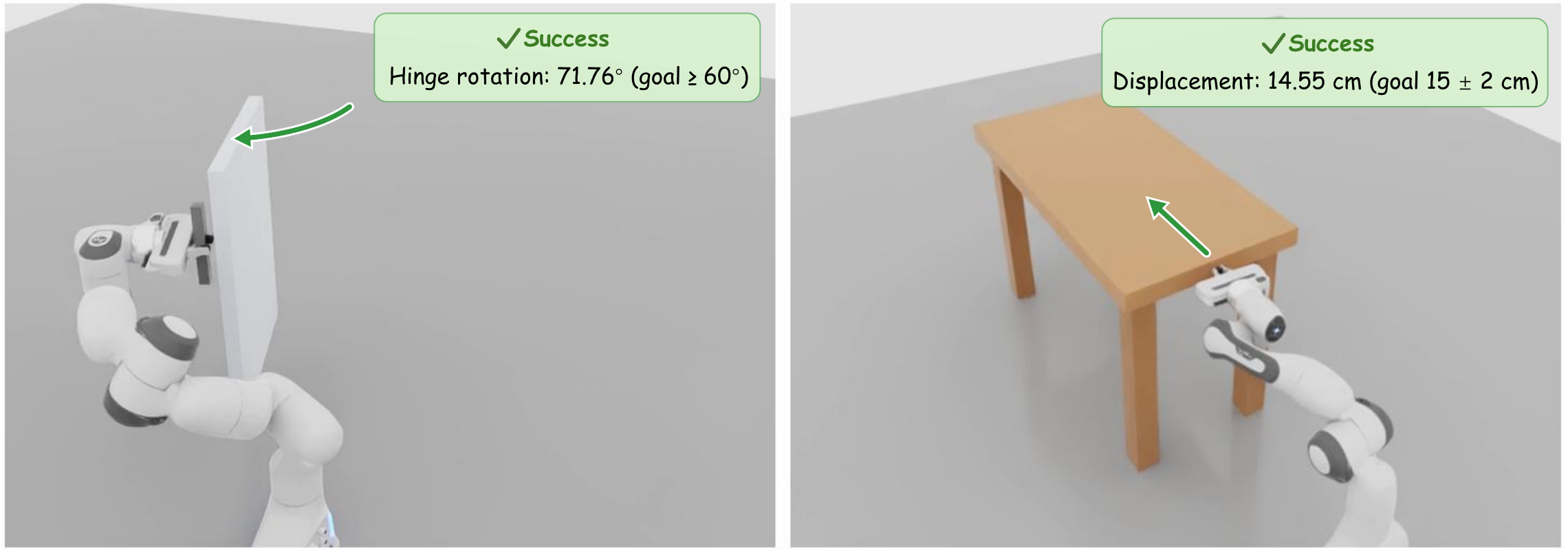}
    \caption{Examples of robot policy calibration in controlled scenes. We validate the Franka Panda execution policies for hinged opening (left) and rigid pushing (right) before evaluating generated scenes.}
   \label{fig:robot_calibration}
\end{figure}

Before the interaction execution, we first calibrate and validate the robot execution policies on positive-control scenes (as shown in~\cref{fig:robot_calibration}), verifying that they can successfully complete the prescribed interactions. We then apply the validated policies to generated scenes, performing target-specific setup, including object binding, contact geometry checks, and robot placement when applicable. This procedure helps distinguish failures arising from the generated scenes' physical properties from those caused by the robot controller. Throughout evaluation, we preserve the generated scenes' native physical properties and the predefined success criteria.

Each interaction starts independently from the original scene state. After execution, the actuator is released or withdrawn, followed by five seconds of post-interaction settling. We record interaction videos at 12 FPS for subsequent scoring. Tasks lacking the required target or native physical capability are marked as not supported (N.S.A.).

\section{Additional Results}
\label{sec:additional_results}
We provide additional quantitative and qualitative results in ScIntBench.

\subsection{Quantitative Results}
\Cref{tab:complete_evaluation} reports the complete results of all baselines, our method, and its ablations, together with the average number of objects per scene (\#Obj). \Cref{tab:overall_efficiency} further reports the computational cost of all methods, including runtime, API cost, and the number of repair candidates per scene.

\begin{figure*}[t]
    \centering
    \includegraphics[width=\linewidth]{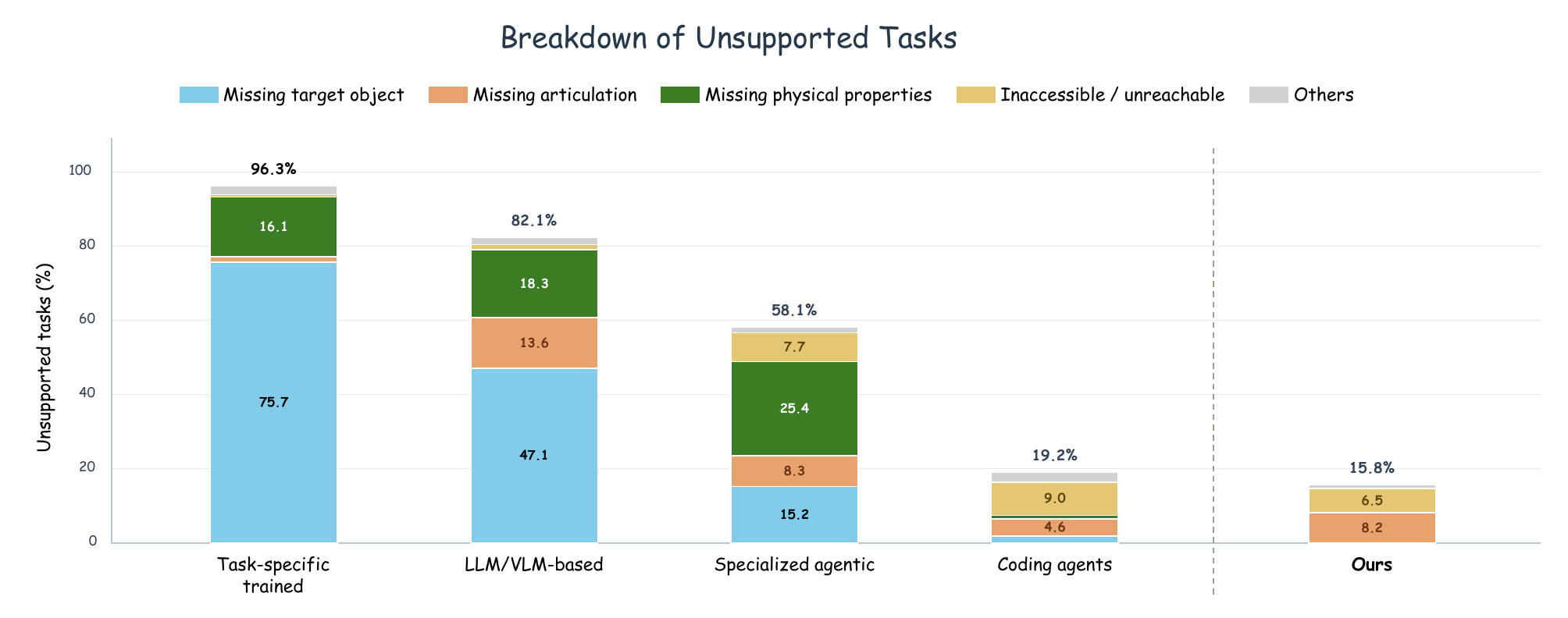}
    \caption{\textbf{Breakdown of unsupported tasks across different scene generation method categories and our method.}}
    \label{fig:nsa_breakdown}
\end{figure*}

To understand why tasks are not executed, \Cref{fig:nsa_breakdown} breaks down the unsupported (N.S.A.) tasks by reason for each method category and our method. The reasons include missing target objects, missing articulation, missing physical properties, inaccessibility, and others. Each unsupported task is assigned to the first unmet reason in this order, since the presence of the target object precedes its articulation and physical properties. The major problems are missing targets and physical properties, while our method achieves the lowest unsupported task rate and almost eliminates both.

\begin{table*}[t]
\centering
\caption{\textbf{Performance comparison of different scene generation methods on ScIntBench and our method with its ablations.} Metric definitions are the same as in \cref{tab:overall_evaluation}. \#Obj denotes the average number of objects per scene for each method. \textbf{Bold} = best, \underline{underline} = second best.}
\label{tab:complete_evaluation}
\setlength{\tabcolsep}{3pt}
\renewcommand{\arraystretch}{1.15}
\resizebox{\linewidth}{!}{%
\begin{tabular}{lc*{8}{c}>{\color{gray}}c*{4}{c}}
\toprule
\multirow{2}{*}{Method} &
\multirow{2}{*}{\#Obj} &
\multicolumn{4}{c}{Semantic Quality} &
\multicolumn{4}{c}{Static Physical Plausibility} &
\multicolumn{5}{c}{Interactive Physical Plausibility} \\
\cmidrule(lr){3-6}
\cmidrule(lr){7-10}
\cmidrule(lr){11-15}
& & 
Real $\uparrow$ &
Func $\uparrow$ &
Lay $\uparrow$ &
Comp $\uparrow$ &
COL $\downarrow$ &
OB $\downarrow$ &
SUP $\uparrow$ &
STAB $\uparrow$ &
\multicolumn{1}{c}{EXE} $\uparrow$&
ISR $\uparrow$ &
ICaP $\uparrow$ &
PIV $\downarrow$ &
DP $\uparrow$ \\
\midrule

\multicolumn{15}{l}{
    \textit{Task-specific generators}} \\

ATISS
    & 10
    & \mstd{3.00}{0.99}
    & \mstd{1.20}{1.34}
    & \mstd{1.92}{1.03}
    & \mstd{1.78}{0.82}
    & \mstd{60.88}{22.38}
    & \mstd{27.45}{14.57}
    & \mstd{27.89}{16.35}
    & N.S.A.
    & 0/584
    & \mstd{0.00}{0.00}
    & \mstd{0.45}{0.57}
    & N.S.A.
    & N.S.A. \\

DiffuScene
    & 11
    & \mstd{3.85}{1.04}
    & \mstd{1.10}{1.21}
    & \mstd{2.50}{0.83}
    & \mstd{2.00}{0.73}
    & \mstd{39.15}{23.06}
    & \mstd{15.56}{11.99}
    & \mstd{66.36}{19.28}
    & N.S.A.
    & 0/584
    & \mstd{0.00}{0.00}
    & \mstd{0.34}{0.42}
    & N.S.A.
    & N.S.A. \\

InstructScene
    & 9
    & \mstd{3.00}{0.65}
    & \mstd{1.05}{1.36}
    & \mstd{1.95}{0.76}
    & \mstd{1.65}{0.67}
    & \mstd{35.11}{22.30}
    & \mstd{19.04}{16.77}
    & \mstd{63.68}{17.28}
    & N.S.A.
    & 0/584
    & \mstd{0.00}{0.00}
    & \mstd{0.33}{0.47}
    & N.S.A.
    & N.S.A. \\

PhyScene
    & 11
    & \mstd{2.56}{0.95}
    & \mstd{1.20}{1.21}
    & \mstd{1.82}{0.87}
    & \mstd{1.46}{0.81}
    & \mstd{41.06}{21.98}
    & \mstd{39.04}{16.75}
    & \mstd{22.55}{14.86}
    & \mstd{94.66}{11.77}
    & 86/584
    & \mstd{4.89}{6.59}
    & \mstd{1.20}{1.47}
    & \underline{\mstd{2.60}{6.76}}
    & \mstd{5.40}{0.27} \\

\midrule
\multicolumn{15}{l}{
    \textit{LLM/VLM-based generators}} \\

LayoutGPT
    & 7
    & \mstd{2.52}{1.05}
    & \mstd{3.22}{1.40}
    & \mstd{2.16}{1.18}
    & \mstd{1.70}{0.81}
    & \mstd{29.14}{30.58}
    & \mstd{41.37}{28.11}
    & \mstd{9.66}{16.04}
    & N.S.A.
    & 0/584
    & \mstd{0.00}{0.00}
    & \mstd{1.39}{0.67}
    & N.S.A.
    & N.S.A.\\

LayoutVLM
    & 9
    & \mstd{3.04}{0.95}
    & \mstd{3.94}{1.19}
    & \mstd{3.04}{1.26}
    & \mstd{2.10}{0.74}
    & \mstd{33.63}{27.65}
    & \mstd{17.78}{19.37}
    & \mstd{76.20}{21.48}
    & N.S.A.
    & 0/584
    & \mstd{0.00}{0.00}
    & \mstd{1.66}{0.49}
    & N.S.A.
    & N.S.A.\\

Holodeck
    & 20
    & \mstd{4.60}{1.14}
    & \mstd{3.22}{1.57}
    & \mstd{2.64}{0.83}
    & \mstd{2.64}{1.24}
    & \mstd{13.48}{13.38}
    & \mstd{4.95}{6.16}
    & \mstd{33.61}{15.62}
    & \mstd{54.25}{41.58}
    & 192/584
    & \mstd{9.41}{10.52}
    & \mstd{2.58}{1.58}
    & \mstd{5.95}{8.49}
    & \mstd{5.89}{1.02}\\

HSM
    & 22
    & \mstd{4.20}{0.93}
    & \mstd{4.35}{1.33}
    & \mstd{3.61}{1.10}
    & \mstd{3.55}{0.91}
    & \mstd{16.30}{14.47}
    & \mstd{10.36}{8.91}
    & \mstd{73.64}{15.73}
    & N.S.A.
    & 0/584
    & \mstd{0.00}{0.00}
    & \mstd{1.62}{0.48}
    & N.S.A.
    & N.S.A.\\

PAT3D
    & 12
    & \mstd{2.53}{1.01}
    & \mstd{3.60}{1.25}
    & \mstd{2.00}{1.29}
    & \mstd{2.37}{0.61}
    & \mstd{28.60}{23.64}
    & \mstd{46.89}{27.04}
    & \mstd{38.38}{25.81}
    & \mstd{60.79}{28.92}
    & 330/584
    & \mstd{15.63}{15.54}
    & \mstd{4.77}{1.52}
    & \mstd{29.28}{15.93}
    & \mstd{5.61}{0.98} \\

\midrule
\multicolumn{15}{l}{
    \textit{Specialized agentic systems}} \\

SceneWeaver
    & 9
    & \mstd{3.60}{1.63}
    & \mstd{3.93}{1.78}
    & \mstd{3.87}{1.93}
    & \mstd{2.67}{1.30}
    & \mstd{12.36}{20.02}
    & \mstd{5.14}{11.96}
    & \mstd{8.58}{13.64}
    & N.S.A.
    & 0/584
    & \mstd{0.00}{0.00}
    & \mstd{1.40}{0.63}
    & N.S.A.
    & N.S.A.\\

SAGE
    & 48
    & \mstd{4.88}{1.33}
    & \mstd{4.46}{1.63}
    & \mstd{3.02}{1.24}
    & \underline{\mstd{4.74}{1.98}}
    & \mstd{0.67}{1.40}
    & \textbf{\mstd{0.00}{0.00}}
    & \mstd{76.63}{12.46}
    & \mstd{98.38}{3.31}
    & 315/584
    & \mstd{6.35}{8.92}
    & \mstd{4.67}{1.57}
    & \textbf{\mstd{0.97}{1.30}}
    & \mstd{6.31}{0.91}\\

SceneSmith
    & 41
    & \underline{\mstd{6.80}{1.34}}
    & \mstd{6.78}{1.88}
    & \mstd{5.80}{1.55}
    & \textbf{\mstd{5.26}{1.65}}
    & \mstd{8.35}{8.01}
    & \mstd{4.67}{4.58}
    & \mstd{94.44}{7.07}
    & \mstd{72.32}{26.14}
    & 419/584
    & \mstd{21.64}{18.38}
    & \mstd{5.81}{1.45}
    & \mstd{6.29}{6.61}
    & \mstd{6.72}{0.73}\\

\midrule
\multicolumn{15}{l}{
    \textit{General-purpose coding agents}} \\

Codex
    & 18
    & \mstd{6.74}{0.80}
    & \underline{\mstd{8.36}{1.06}}
    & \underline{\mstd{7.40}{1.14}}
    & \mstd{4.46}{1.30}
    & \mstd{3.48}{10.29}
    & \mstd{4.92}{9.51}
    & \mstd{85.42}{19.05}
    & \mstd{97.75}{6.85}
    & 470/584
    & \mstd{31.93}{14.74}
    & \mstd{6.58}{0.95}
    & \mstd{7.70}{9.71}
    & \underline{\mstd{6.77}{0.51}}\\

Claude Code
    & 24
    & \textbf{\mstd{7.08}{0.80}}
    & \textbf{\mstd{8.62}{1.09}}
    & \textbf{\mstd{7.62}{0.81}}
    & \mstd{4.70}{1.28}
    & \mstd{1.06}{3.68}
    & \mstd{0.40}{2.83}
    & \mstd{94.05}{8.88}
    & \mstd{92.30}{21.15}
    & 474/584
    & \mstd{36.99}{20.21}
    & \underline{\mstd{7.30}{1.07}}
    & \mstd{15.03}{14.73}
    & \mstd{6.63}{1.16}\\
\midrule
\multicolumn{15}{l}{
    \textit{Interaction-aware generation methods}} \\
w/o Static Critic
    & 16
    & \mstd{6.33}{1.05}
    & \mstd{7.77}{1.45}
    & \mstd{7.20}{0.85}
    & \mstd{3.97}{1.30}
    & \mstd{0.95}{5.22}
    & \mstd{0.18}{0.96}
    & \mstd{96.79}{7.19}
    & \mstd{97.44}{3.04}
    & 483/584
    & \mstd{37.46}{21.46}
    & \mstd{7.17}{1.39}
    & \mstd{5.77}{6.15}
    & \mstd{6.18}{1.28}\\

w/o Dynamic Critic
    & 16
    & \mstd{6.27}{0.92}
    & \mstd{7.67}{1.56}
    & \mstd{6.87}{1.11}
    & \mstd{3.87}{0.90}
    & \textbf{\mstd{0.00}{0.00}}
    & \mstd{0.32}{1.92}
    & \mstd{97.01}{5.08}
    & \mstd{99.43}{3.09}
    & 476/584
    & \mstd{35.73}{21.08}
    & \mstd{7.26}{1.34}
    & \mstd{7.57}{9.73}
    & \mstd{6.70}{0.63}\\

w/o Diagnosis
    & 17
    & \mstd{6.40}{0.71}
    & \mstd{7.73}{1.55}
    & \mstd{7.33}{0.84}
    & \mstd{4.00}{1.05}
    & \textbf{\mstd{0.00}{0.00}}
    & \textbf{\mstd{0.00}{0.00}}
    & \underline{\mstd{97.21}{5.93}}
    & \textbf{\mstd{100.00}{0.00}}
    & 480/584
    & \underline{\mstd{38.54}{18.31}}
    & \mstd{7.20}{1.31}
    & \mstd{6.15}{8.37}
    & \mstd{6.58}{0.43}\\

\textbf{Ours (Full)}
    & 16
    & \mstd{6.43}{0.99}
    & \mstd{7.80}{1.45}
    & \mstd{7.17}{1.15}
    & \mstd{3.90}{0.99}
    & \textbf{\mstd{0.00}{0.00}}
    & \mstd{0.35}{1.92}
    & \textbf{\mstd{97.26}{7.00}}
    & \textbf{\mstd{100.00}{0.00}}
    & 492/584
    & \textbf{\mstd{43.03}{16.76}}
    & \textbf{\mstd{7.52}{1.21}}
    & \mstd{3.34}{5.25}
    & \textbf{\mstd{6.83}{0.64}}\\
\bottomrule
\end{tabular}%
}

\end{table*}

\begin{table*}[t]
\centering
\caption{\textbf{Computational cost per scene, measured by runtime and API cost.} $^\dagger$API cost estimated from token usage at official pricing. We also report the number of repair candidates explored per scene.}
\label{tab:overall_efficiency}
\centering
\scriptsize
\setlength{\tabcolsep}{4pt}
\begin{tabular*}{\textwidth}{@{\extracolsep{\fill}}lrrrrrrc@{}}
    \toprule
    \multirow{2}{*}{Method}
    & \multicolumn{3}{c}{Runtime (min)}
    & \multicolumn{3}{c}{API Cost (USD)} 
    & \multirow{2}{*}{Repair Candidates}\\
    \cmidrule(lr){2-4}
    \cmidrule(l){5-7}
    & Mean & Min & Max & Mean & Min & Max &\\
    \midrule
    ATISS & 0.27 & 0.15 & 0.72 & 0 & 0 & 0 & --\\
    DiffuScene & 1.08 & 0.51 & 1.29 & 0 & 0 & 0 & --\\
    InstructScene & 0.26 & 0.03 & 0.66 & 0 & 0 & 0 & --\\
    PhyScene & 2.09 & 1.27 & 3.08 & 0 & 0 & 0 & --\\
    \midrule
    LayoutGPT & 0.41 & 0.10 & 1.98 & 0.13 & 0.08 & 0.55 & --\\
    LayoutVLM & 1.41 & 0.66 & 3.49 & 0.15 & 0.10 & 0.47 & --\\
    Holodeck & 3.61 & 2.36 & 4.82 & 0.05 & 0.03 & 0.08 & --\\
    HSM & 9.04 & 3.53 & 20.98 & 0.64 & 0.22 & 1.90 & 7.2\\
    PAT3D & 46.73 & 14.21 & 88.03 & 1.12 & 0.28 & 2.83 & 2.3\\
    \midrule
    SceneWeaver & 114.09 & 5.61 & 209.63 & 0.64 & 0.09 & 2.01 & 3.4\\
    SAGE & 113.99 & 36.68 & 370.00 & 0.40 & 0.17 & 0.99 & 3.8\\
    SceneSmith & 148.33 & 56.05 & 455.42 & 11.42 & 2.97 & 24.77 & 18.9\\
    \midrule
    Codex & 25.66 & 15.45 & 36.66 
    & 6.54$^\dagger$ & 4.32$^\dagger$ & 10.05$^\dagger$ & 4.6\\
    Claude Code & 31.39 & 16.28 & 59.93 
    & 4.65$^\dagger$ & 3.06$^\dagger$ & 6.91$^\dagger$ & 1.7\\
    \midrule
    w/o Static Critic & 115.29 & 38.57 & 242.23 & 11.88$^\dagger$ 
    & 3.35$^\dagger$ & 17.34$^\dagger$ & 10.7\\
    w/o Dynamic Critic & 19.46 & 5.12 & 57.74 & 4.66$^\dagger$ 
    & 1.33$^\dagger$ & 11.36$^\dagger$ & 2.8\\
    w/o Diagnosis & 128.22 & 27.07 & 289.20 & 12.59$^\dagger$ 
    & 3.43$^\dagger$ & 20.23$^\dagger$ & 14.7\\
    Ours (Full) & 119.35 & 28.10 & 280.63 & 10.33$^\dagger$ 
    & 5.01$^\dagger$ & 18.15$^\dagger$ & 11.8\\
    \bottomrule
\end{tabular*}
\end{table*}

\begin{table*}[t]
\centering
\caption{\textbf{Verification of geometry and physical properties during conversion to Isaac Sim.} We randomly select ten scenes for each method. Max Mesh Error is the largest absolute difference in any coordinate of corresponding source and USD mesh vertices. We count comparable property records traced to the original outputs or their loaders. For each scene, we compute the percentage of these records that match the USD declarations and report the mean value across scenes. After loading each scene in Isaac Sim, we verify rigid-body masses, material bindings, friction, restitution, and joints in PhysX. We report the number of checks passed out of the total tested across the ten scenes. \textit{N/A}: the property is not provided by the method's output, so there is nothing to verify.}
\label{tab:conversion_audit}
\setlength{\tabcolsep}{3pt}
\renewcommand{\arraystretch}{1.15}
\resizebox{\linewidth}{!}{%

\begin{tabular}{@{}lrrrrrrrr@{}}
\toprule
& \multicolumn{3}{c}{Source-to-USD Checks}
& \multicolumn{5}{c}{Isaac Sim Runtime Checks (passed/tested)} \\
\cmidrule(lr){2-4}\cmidrule(lr){5-9}
Method & Max Mesh Error ($\mu$m) & \# Traced Properties
& Property Match Rate (\%)
& Mass & Material Bindings & Friction & Restitution & Joints \\
\midrule

ATISS
& 0.180 & 0
& \textit{N/A} & \textit{N/A} & \textit{N/A}
& \textit{N/A} & \textit{N/A} & \textit{N/A}\\

DiffuScene
& 0.000 & 0
& \textit{N/A} & \textit{N/A} & \textit{N/A}
& \textit{N/A} & \textit{N/A} & \textit{N/A}\\

InstructScene
& 0.000 & 0
& \textit{N/A} & \textit{N/A} & \textit{N/A}
& \textit{N/A} & \textit{N/A} & \textit{N/A}\\

LayoutGPT
& 0.119 & 0
& \textit{N/A} & \textit{N/A} & \textit{N/A}
& \textit{N/A} & \textit{N/A} & \textit{N/A}\\

LayoutVLM
& 0.243 & 0
& \textit{N/A} & \textit{N/A} & \textit{N/A}
& \textit{N/A} & \textit{N/A} & \textit{N/A}\\

HSM
& 0.237 & 0
& \textit{N/A} & \textit{N/A} & \textit{N/A}
& \textit{N/A} & \textit{N/A} & \textit{N/A}\\

SceneWeaver
& 0.295 & 0
& \textit{N/A} & \textit{N/A} & \textit{N/A}
& \textit{N/A} & \textit{N/A} & \textit{N/A}\\

PhyScene
& 0.109 & 520
& 100.00
& 168/168 & 2,886/2,886
& 168/168 & 168/168 & 130/130\\

Holodeck
& 1.676 & 492
& 100.00
& 214/214 & 40/40
& 214/214 & 214/214 & \textit{N/A}\\

PAT3D
& 1.331 & 132
& 100.00
& 62/62 & 72/72
& 62/62 & 62/62 &\textit{N/A}\\

SAGE
& 0.287 & 8,833
& 100.00
& 585/585 & 788/788
& 585/585 & 585/585 &\textit{N/A}\\

SceneSmith
& 1.911 & 59,575
& 100.00
& 333/333 & 26,740/26,740
& 333/333 & 333/333 &36/36\\

Codex
& 0.000 & 7,796
& 100.00
& 223/223 & 1,310/1,310
& 223/223 & 223/223 &16/16\\

Claude Code
& 0.000 & 10,316
& 100.00
& 256/256 & 1,562/1,562
& 256/256 & 256/256 &26/26\\
\bottomrule
\end{tabular}
}
\end{table*}

\subsection{Qualitative Results}
\begin{figure*}[t]
    \centering
    \includegraphics[width=\linewidth]{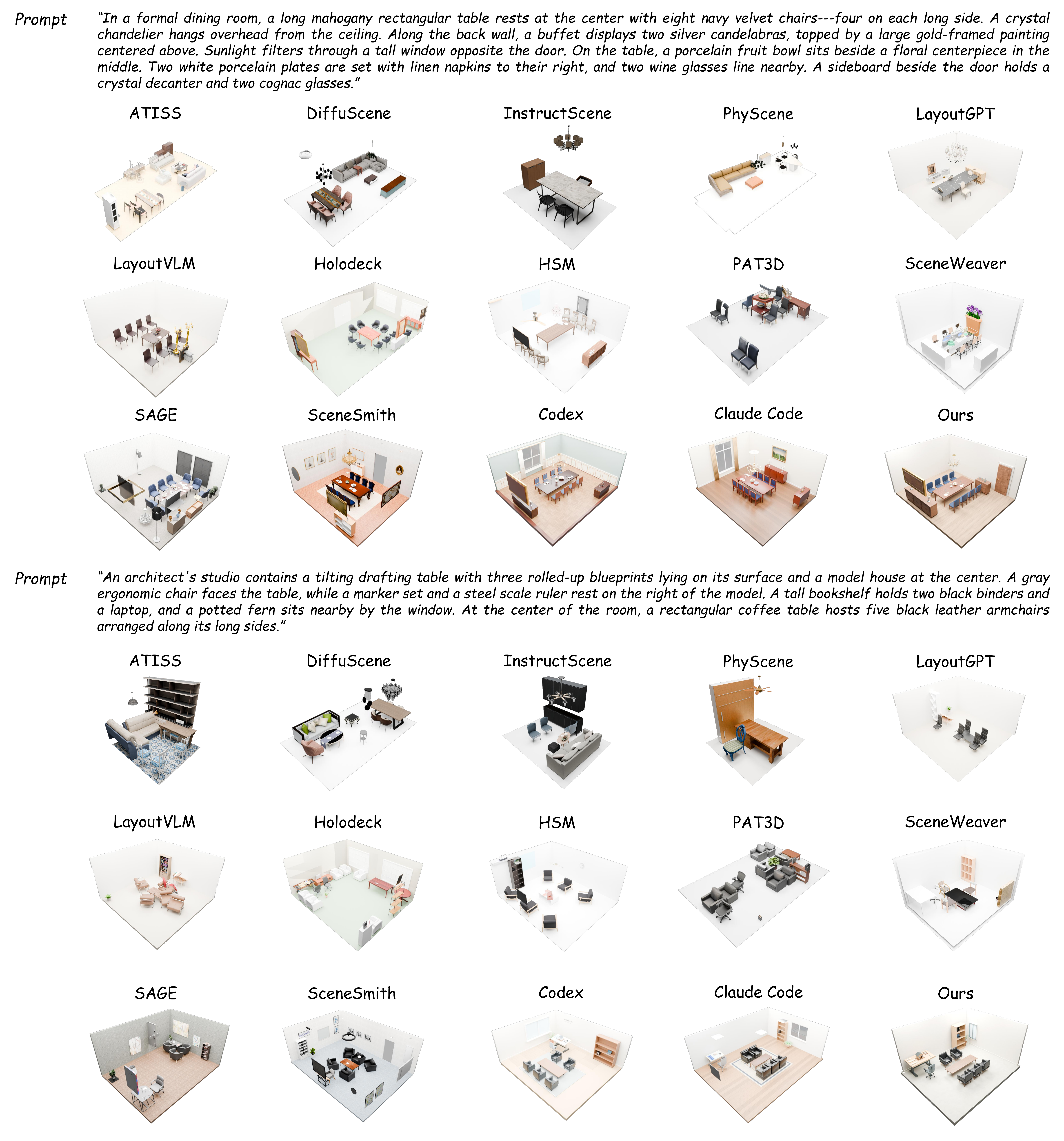}
    \caption{\textbf{Additional qualitative comparison of generated scenes across 15 scene generation methods.} }
    \label{fig:more_qualitative}
\end{figure*}

\begin{figure*}[t]
    \centering
    \includegraphics[width=0.7\linewidth]{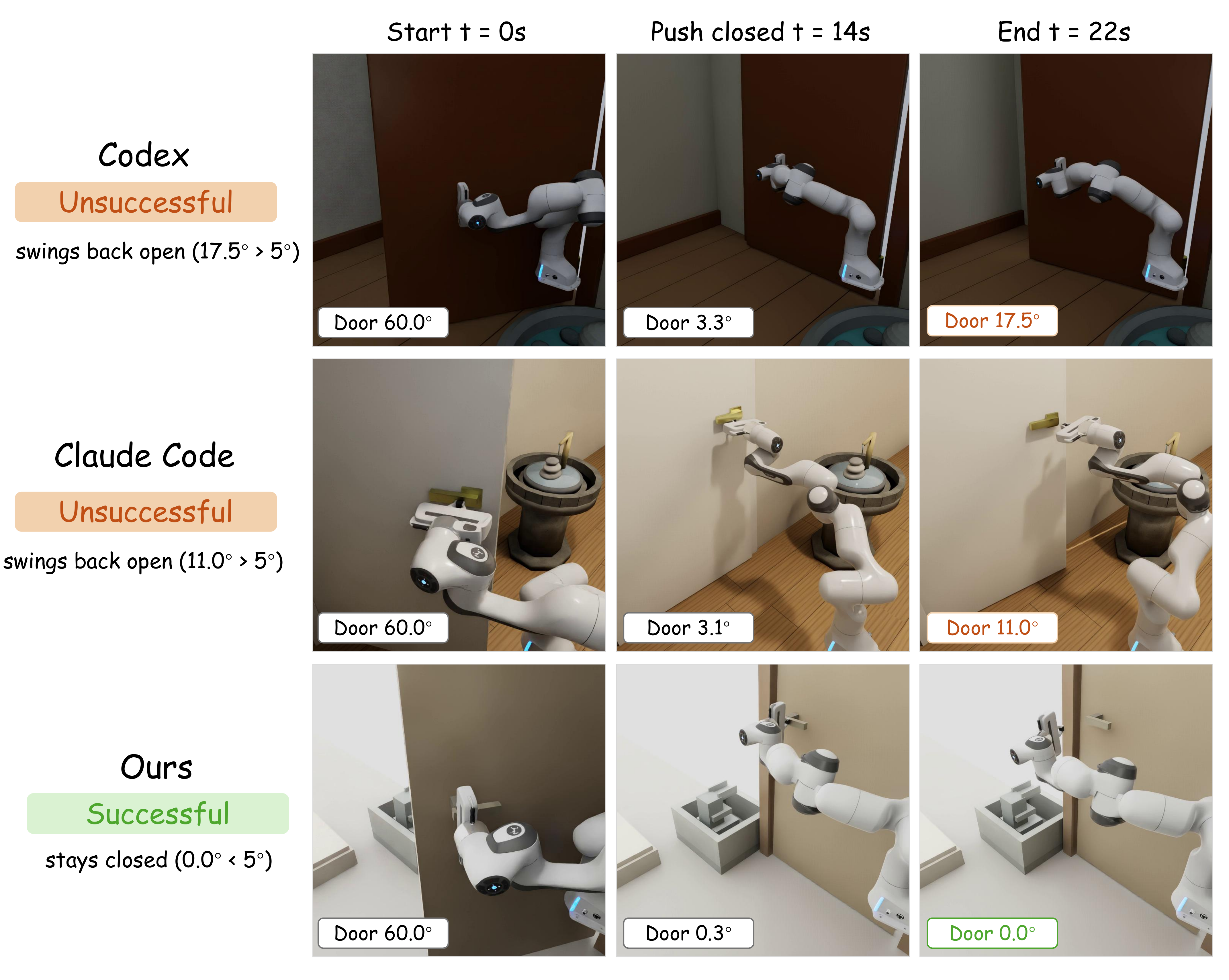}
    \caption{\textbf{Qualitative comparison of interaction outcomes on the \textit{close the door} task.} The doors in the Codex and Claude Code scenes swing back open after being closed (final angle $>5^\circ$), whereas the door in our scene stays closed.}
    \label{fig:qualitative3}
\end{figure*}

\Cref{fig:more_qualitative} presents two more qualitative comparisons of scenes generated by all 15 methods (14 baselines and ours) using two shared text prompts that describe a formal dining room and an architect's studio, respectively. Recent methods, such as SAGE, SceneSmith, and the coding agents, produce visually rich and well-furnished scenes that closely follow the prompts, whereas earlier task-specific generators often yield sparse layouts without room structure. However, such visual quality does not reveal whether a scene supports physical interaction. Furthermore, \Cref{fig:qualitative3} illustrates a qualitative example of the \textit{close the door} task. The doors in the Codex and Claude Code scenes swing back open after being closed, whereas the door in our scene stays closed, indicating that our dynamic critic repairs interaction failures that static checks cannot detect.

\section{Benchmark Validation}
\label{sec:benchmark_validation}
\begin{table*}[t]
\caption{\textbf{Comparison with interactive reference scenes.} We report 19 matched task assignments for reference execution: 9 from BEHAVIOR-1K~\cite{li2023behavior}, 3 from ReplicaCAD~\cite{szot2021habitat}, and 7 from iGibson~\cite{li22b}. $\uparrow$ indicates higher is better, and $\downarrow$ indicates lower is better. \textbf{Bold} = best.}
\label{tab:reference_scene_comparison}
\centering
\begingroup
\footnotesize
\setlength{\tabcolsep}{3pt}
\renewcommand{\arraystretch}{1.08}
\begin{tabular*}{\textwidth}{@{\extracolsep{\fill}}llccccc@{}}
\toprule
Reference set & Source &
EXE$\uparrow$ &
ISR\,(\%)$\uparrow$ &
ICaP$\uparrow$ &
PIV\,(\%)$\downarrow$ &
DP$\uparrow$ \\
\midrule
\textbf{BEHAVIOR-1K} & \textbf{Reference} & 9/9 & \textbf{100.00} & \textbf{10.00} 
& \textbf{0.00} & \textbf{7.78} \\
& SAGE & 5/9 & 11.11 & 4.89 & 0.53 & 6.00 \\
& SceneSmith & 6/9 & 22.22 & 5.11 & 5.62 & 5.83 \\
& Holodeck & 4/9 & 33.33 & 4.67 & 6.58 & 6.50 \\
& Codex & 6/9 & 44.44 & 6.22 & 2.78 & 7.17 \\
& Claude Code & 7/9 & 66.67 & 8.44 & 13.31 & 7.43 \\
\midrule
\textbf{ReplicaCAD} & \textbf{Reference} & 3/3 & \textbf{100.00} & \textbf{10.00} 
& \textbf{0.00} & 7.00 \\
& SAGE & 2/3 & 0.00 & 6.00 & 1.43 & 7.00 \\
& SceneSmith & 2/3 & 66.67 & 7.33 & 10.28 & \textbf{8.00} \\
& Holodeck & 1/3 & 33.33 & 4.67 & \textbf{0.00} & 7.00 \\
& Codex & 1/3 & 33.33 & 4.67 & \textbf{0.00} & 6.50 \\
& Claude Code & 2/3 & 33.33 & 7.33 & 45.19 & 4.50 \\
\midrule
\textbf{iGibson} & \textbf{Reference} & 7/7 & \textbf{100.00} & \textbf{10.00} 
& \textbf{0.00} & 6.57 \\
& SAGE & 2/7 & 14.29 & 3.71 & \textbf{0.00} & 6.00 \\
& SceneSmith & 3/7 & 28.57 & 4.86 & 19.62 & 5.00 \\
& Holodeck & 1/7 & 14.29 & 3.14 & \textbf{0.00} & 7.00 \\
& Codex & 6/7 & 42.86 & 7.71 & \textbf{0.00} & 7.50 \\
& Claude Code & 6/7 & 42.86 & 8.00 & 9.50 & \textbf{8.00} \\
\bottomrule
\end{tabular*}
\endgroup
\end{table*}

\subsection{Validation on Reference Scenes}
\label{sec:reference_scenes}
\Cref{tab:reference_scene_comparison} reports the comparison between selected interactive reference scenes and scenes generated by five methods. We select rooms or regions from BEHAVIOR-1K~\cite{li2023behavior}, ReplicaCAD~\cite{szot2021habitat}, and iGibson~\cite{li22b} based on their similarity to the input scene descriptions and the presence of relevant interaction targets. We evaluate matched tasks using fixed task-specific protocols and success criteria. The comparison includes 9 planned interaction tasks in BEHAVIOR-1K, 3 in ReplicaCAD, and 7 in iGibson, with results aggregated across the matched task assignments within each reference set.

\subsection{Human Evaluation}
\label{sec:human_eval}

\begin{table}[t]
\centering
\caption{\textbf{Validation on LLM-assigned tasks.} Approval rate denotes the averaged Yes judgments, while Agreement is the rate of tasks for which all participant have the same judgment.}
\label{tab:human_task_validation}
\resizebox{\linewidth}{!}{
\begin{tabular}{lcc}
\toprule
\textbf{Dimension} &
\textbf{Approval Rate (\%) $\uparrow$} &
\textbf{Agreement (\%) $\uparrow$}\\
\midrule
Task Relevance & 100.0\% & 100.0\%  \\
Task Feasibility  & 99.7\% & 99.0\% \\
Criterion Validity & 95.3\% & 91.0\% \\
\bottomrule
\end{tabular}
}
\end{table}

As described in the main paper, four annotators participated in both studies.

\paragraph{Ethics.} No ethics review was conducted, because the studies only collected participants' judgments about generated scenes, interaction tasks, and simulated videos, involved no personal or sensitive information, and posed no risk to participants. All participants volunteered, were informed of the study purpose and procedure, and gave consent before participation.

\subsubsection{Interaction Task Validation}
\begin{figure*}[t]
    \centering
    \includegraphics[width=\linewidth]{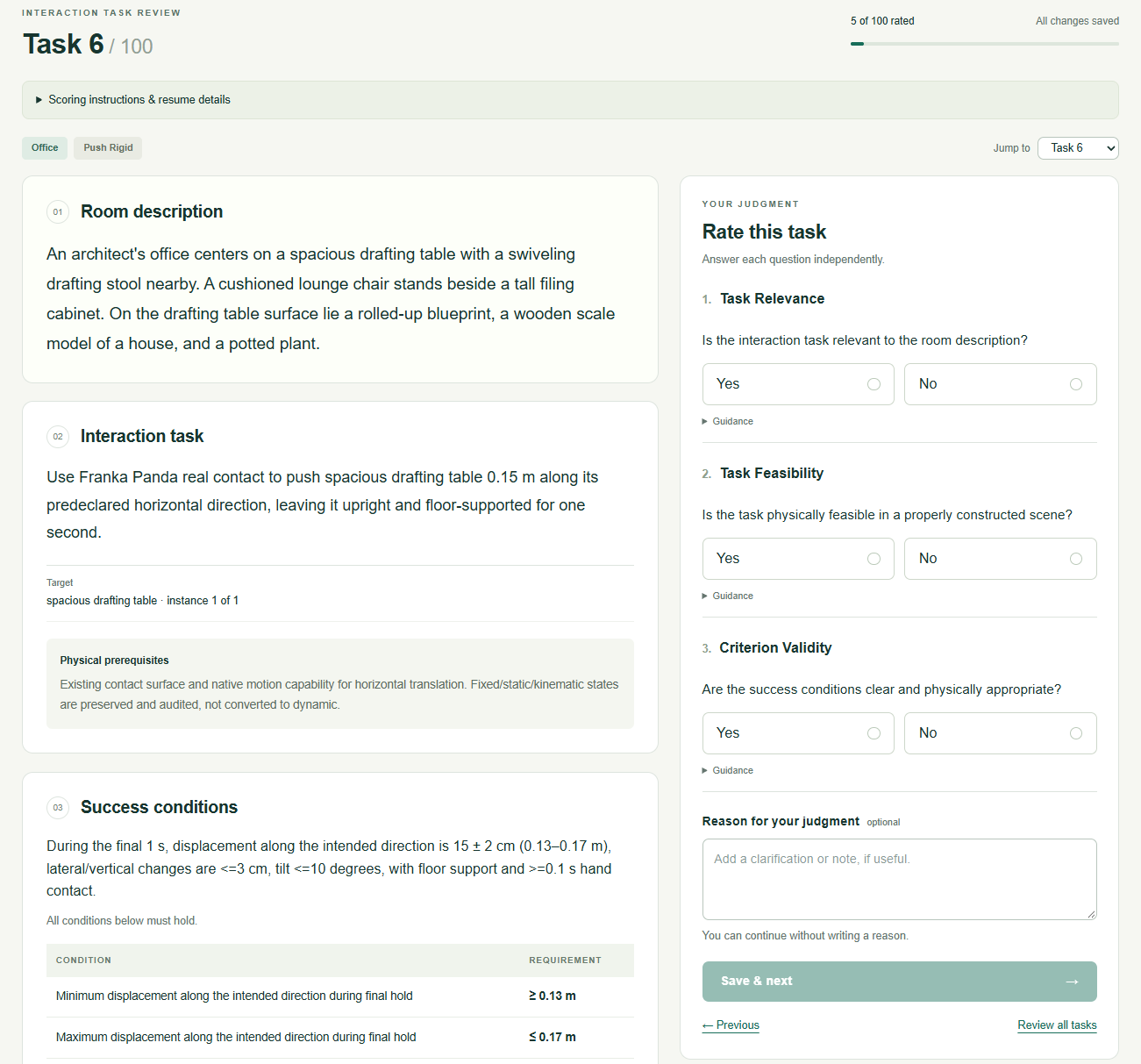}
    \caption{\textbf{Human validation for LLM-assigned tasks.} Participants review the room description, proposed interaction task, and corresponding success conditions. They provide separate Yes/No judgments on task relevance, task physical feasibility, and success criterion validity.}
    \label{fig:task_validation}
\end{figure*}

\Cref{fig:task_validation} demonstrates our human evaluation interface for LLM-assigned tasks. Participants are presented with the room description, proposed interaction task, and corresponding success conditions. On the right side of the interface, participants independently provide Yes/No judgments on three dimensions: (a) \textbf{Task relevance}: whether the proposed interaction is relevant to the room description; (b) \textbf{Task feasibility}: whether the task is physically
feasible in a properly constructed scene consistent with the room description and setup; and (c) \textbf{Criterion validity}: whether the success conditions are clear, measurable, and physically appropriate for the intended interaction. The approval rates and the agreement rates for these three dimensions are reported in \cref{tab:human_task_validation}.

\subsubsection{Active Metrics Validation}
\begin{figure*}[t]
    \centering
    \includegraphics[width=\linewidth]{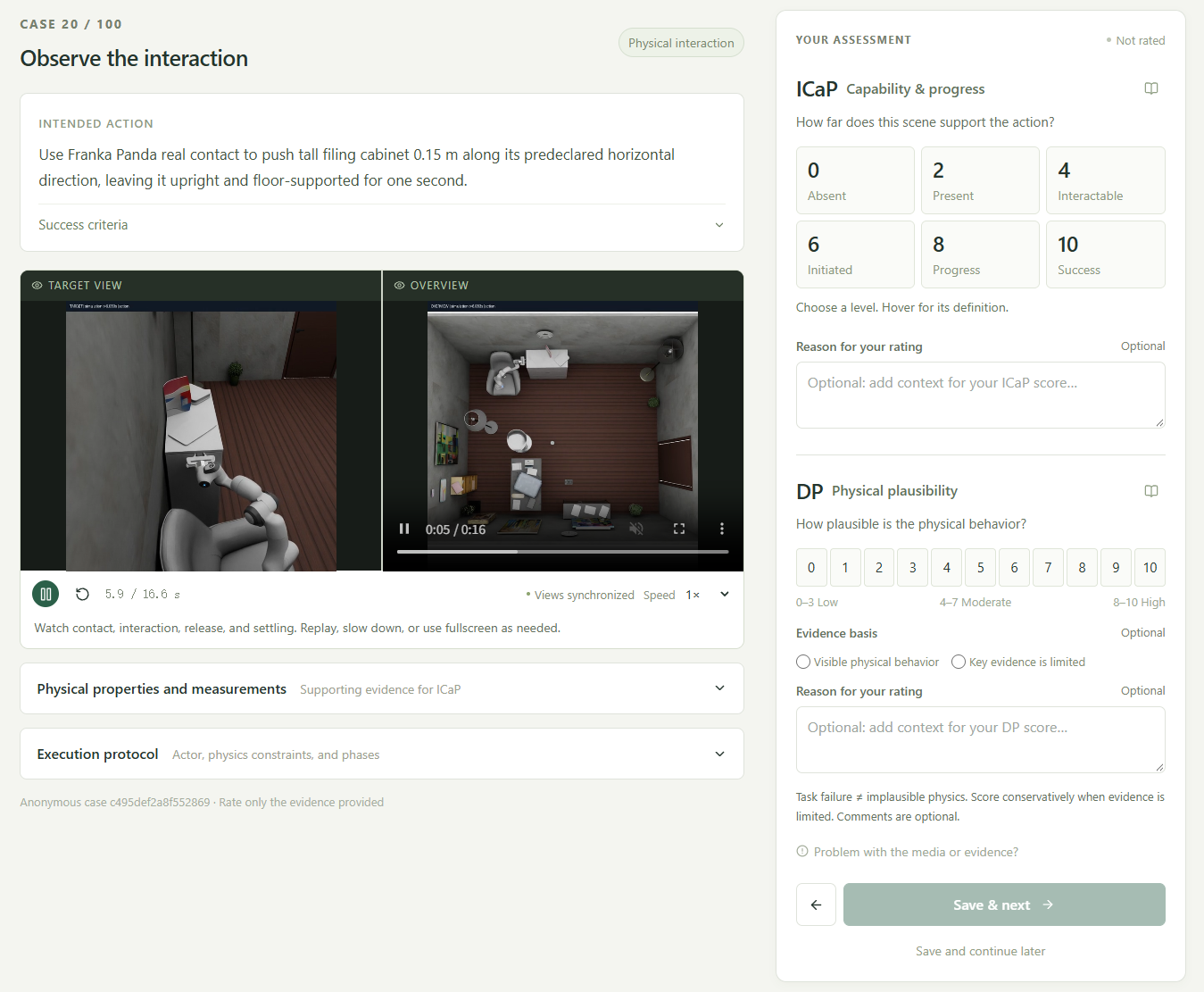}
    \caption{\textbf{Human evaluation interface for active metrics validation.} Participants inspect synchronized target and overview videos, the intended action, and supporting physical measurements. They independently assess interaction capability and progress (ICaP) using a six-level rubric and dynamic plausibility (DP) on a 0--10 scale.}
    \label{fig:human_evaluation}
\end{figure*}

\Cref{fig:human_evaluation} illustrates our human evaluation interface for active metrics. Participants are presented with the intended action and its success criteria, followed by two interaction videos showing the target and overview perspectives. We also provide the physical properties and measurements for reference. On the right side of the interface, participants independently rate interaction capability progress (ICaP) and dynamic plausibility (DP) using the same scoring rubrics as the MLLM-based evaluator in our ScIntBench. Participants may optionally provide reasons for their rating. 

For each metric, we measure the correlation between MLLM scores and mean human ratings using Spearman’s rank correlation coefficient $\rho$ and Kendall’s $\tau$. We also report the human-human correlation. \Cref{tab:human_mllm_agreement} shows that MLLM scores correlate strongly with human ratings. 

\begin{table}[t]
\centering
\caption{\textbf{The correlation between MLLM-based evaluations and human evaluations.} Spearman's $\rho$ and Kendall's $\tau$ are used.}
\label{tab:human_mllm_agreement}
\resizebox{\linewidth}{!}{
\begin{tabular}{lccc}
\toprule
\textbf{Metric} &
\textbf{MLLM-Human $\rho$ $\uparrow$} &
\textbf{MLLM-Human $\tau$ $\uparrow$} &
\textbf{Human-Human $\rho$ $\uparrow$}\\
\midrule
ICaP & 0.818 & 0.749 & 0.848\\
DP  & 0.683 & 0.638 & 0.786\\
\bottomrule
\end{tabular}
}
\end{table}

\end{document}